\documentclass[]{spie}  %>>> use for US letter paper
\usepackage{amsmath,amsfonts,amssymb}
\usepackage{graphicx}
\usepackage[colorlinks=true, allcolors=blue]{hyperref}
\usepackage{epstopdf}
\usepackage{subfigure}
\usepackage{algorithmic}
\usepackage{algorithm}
\DeclareMathOperator*{\argmin}{arg\,min}

\title{Instance Influence Estimation for Hyperspectral Target Signature Characterization using Extended Functions of Multiple Instances}

\author{Sheng Zou}
\author{Alina Zare}
\affil{Department of Electrical and Computer Engineering, University of Missouri, Columbia, USA}

\authorinfo{Further author information:  Alina Zare: zarea@missouri.edu (corresponding author),  Sheng Zou: sz9nf@mail.missouri.edu\\
The authors wish to thank the National Geospatial-Intelligence Agency for support of this research under the project entitled ``NIP: Functions of Multiple Instances for Hyperspectral Analysis.''}

\begin{document} 
\maketitle

\begin{abstract}
The Extended Functions of Multiple Instances ($e$FUMI) algorithm \cite {jiao2015functions} is a generalization of Multiple Instance Learning (MIL). In $e$FUMI, only bag level (i.e. set level) labels are needed to estimate target signatures from mixed data.  The training bags in $e$FUMI are labeled positive if any data point in a bag contains or represents any proportion of the target signature and are labeled as a negative bag if all data points in the bag do not represent any target. From these imprecise labels, $e$FUMI has been shown to be effective at estimating target signatures in hyperspectral subpixel target detection problems.   One motivating scenario for the use of $e$FUMI is where an analyst circles objects/regions of interest in a hyperspectral scene such that the target signatures of these objects can be estimated and be used to determine whether other instances of the object appear elsewhere in the image collection.  The regions highlighted by the analyst serve as the imprecise labels for $e$FUMI.  Often, an analyst may want to iteratively refine their imprecise labels.  In this paper, we present an approach for estimating the influence on the estimated target signature if the label for a particular input data point is modified.  This “instance influence estimation” guides an analyst to focus on (re-)labeling the data points that provide the largest change in the resulting estimated target signature and, thus, reduce the amount of time an analyst needs to spend refining the labels for a hyperspectral scene. Results are shown on real hyperspectral sub-pixel target detection data sets.   
\end{abstract}

% Include a list of keywords after the abstract 

\section{INTRODUCTION}
\label{sec:intro}  % \label{} allows reference to this section

The Extended Functions of Multiple Instance ($e$FUMI) algorithm is a hyperspectral target characterization method that estimates the spectral signature of a target of interest from training data with imprecise and uncertain labels. The $e$FUMI algorithm addresses problems that are generalizations of those addressed by a standard Multiple Instance Learning (MIL) framework \cite {maron1998framework}.  In MIL,  the training points are grouped into ``bags.''  A bag is a multi-set of training data points. Each bag is labeled as ``positive'' or ``negative.''  Under MIL, positive bags are assumed to contains at least one target data point and negative bags contains exclusively non-target points. Generally, MIL algorithms assume that each data point is either entirely a target or a non-target data point.  The $e$FUMI algorithm extends the MIL framework to address data points that are mixtures of target and non-target signatures.  Thus, in $e$FUMI, positive bags are assumed to contain at least one data point that has some non-zero proportion of target.  Negative bags in $e$FUMI are assumed to contain data points that all have zero target proportion.  

The advantage of MIL and $e$FUMI during hyperspectral image analysis is that they allow for bag-level (instead of precise instance-level) training labels.  Creating bag level labels is generally much less time consuming and, often, much more feasible than assigning accurate data-point level training labels.   For example, in the case of sub-pixel target detection in remotely-sensed hyperspectral imagery, GPS coordinate of targets in training data may be known but the accuray of the GPS coordinates may range across several pixels.  Thus, the ground truth in this case is inherently imprecise.  Yet, it is feasible from this imprecise groundtruth data to create positive bags associated with each ground truth point and include all pixels in a neighborhood based on the GPS unit accuracy and, also, create negative bags which are far from all GPS target ground truth points.  

 Another scenario in which $e$FUMI may be useful is one in which an analyst circles regions of interest in an image (and, thus, these regions become positive bags) and also circles regions of no-interest (these become negative bags).  Then, $e$FUMI can take these regions and produce an estimate of the target spectral signature that is found in all positive bags (and not found in any negative bags).  By being able to circle general regions of interest using some sort of user interface, it is much more intuitive and easier for an analyst to generate label maps as opposed to being required to create some sort of pixel level label maps.

 However, differences in labeling may cause in differences in the $e$FUMI result. Thus, after obtaining an $e$FUMI result, an analyst may wish to refine their bags and bag-level labels.  In this paper, we propose two methods for identifying which pixels (or image regions) would cause the largest change in $e$FUMI results if relabeled.  Our two proposed influence estimation methods estimate the relative degree that the resulting $e$FUMI-estimated target endmember will change after changing the labels of the respective pixels/regions.

\section{Extended Functions of Multiple Instances}

The $e$FUMI algorithm is a method to learn a target signature from training data without the need of instance-level labels.\cite {jiao2015functions} Specifically, the goal of $e$FUMI is to estimate a target signature, $\mathbf{e}_T$, non-target signatures, $\left\{ \mathbf{e}_k\right\}_{k=1}^M$, the number of needed non-target signatures, $M$,  and the function parameters, $\mathbf{p}_i$, which define the relationship between each data point, $\mathbf{x}_i$, and the endmember signatures. These are estimated given a set of input training data, $\{\mathbf{x}_i\}_{i=1}^N \in \mathbb{R}^D$ which have been partitioned into $K$ ``bags,'' $\mathbf{B} = \left\{ B_1, \ldots, B_K\right\}$, with associated bag-level labels, $L = \left\{L_1, \ldots, L_K\right\}$.  Each training point is assumed to be a function of target and non-target signatures, $\mathbf{x}_i = f(\mathbf{E}_i, \mathbf{p}_i)$.  In the results shown here, we consider the case of a convex combination for the functional form, $\mathbf{x}_i = p_{iT}\mathbf{e}_T + \sum_{k=1}^M p_{ik}\mathbf{e}_k$ subject to the constraints that $p_{iT} + \sum_{k=1}^M p_{ik} = 1$, $p_{iT} \ge 0$, $ p_{ik} \ge 0$, $\left\|\mathbf{e}_k\right\|^2 = 1$, and $\left\|\mathbf{e}_T\right\|^2 = 1$. Furthermore,  the bag-level labels are \emph{unspecific} since, if for bag $B_j$ with $L_j = 1$ (thus, $B_j$ is a positive bag), this indicates that there is at least one data point in $B_j$ with a positive $p_{iT}$ indicating some presence of target, as in \eqref{eq:l1}, where $\boldsymbol{\varepsilon}_i$ is an error term. 
\vspace{-2mm} 
\begin{equation}
\text{if }L_j = 1, \exists \mathbf{x}_i \in B_j\text{ s.t. }\mathbf{x}_i = p_{iT}\mathbf{e}_T + \sum_{k=1}^{M} p_{ik}\mathbf{e}_{k}+\boldsymbol{\varepsilon}_{i}, p_{iT} > 0 
\label{eq:l1}
\end{equation}
However, the exact number of data points in a positive bag with a target contribution (i.e., $p_{iT} > 0$), is unknown, also, the target proportion is unknown.  
Furthermore, if $B_j$ is a negative bag (i.e., $L_j = 0$), then this indicates that none of the data in $B_j$ contain any target, as in \eqref{eq:l2}.
\begin{equation}
\text{if }L_j = 0,  \forall \mathbf{x}_i \in B_j, \mathbf{x}_i =  \sum_{k=1}^{M} p_{ik}\mathbf{e}_{k}+\boldsymbol{\varepsilon}_{i}
 \label{eq:l2} \vspace{-2mm}
\end{equation}

Given training data of this form, instance-level labels are unknown. $e$FUMI addresses this problem using an Expectation Maximization (EM) approach in which the instance-level labels are the hidden, latent variables in the EM model. The assumed \emph{complete} data log-likelihood is proportional to a cost function \cite {zare2014extended} with an approximation error term (minimizing the L2 norm between training points and $e$FUMI-estimated reconstructed per-point signatures),  endmember constraint terms (minimizing the L2 norm between endmembers and global data mean) and a sparsity promoting term (promoting the proportions associated with unnecessary endmembers to be zero). The optimization of the cost function is solved by iteratively via EM. Since instance-level labels are unknown for each training point, in the E-step, the conditional probability of an instance-level label given parameters and bag-level label is computed. In the M-step,  estimated endmembers and proportions are updated by minimizing the expected value of cost function with respect to the latent instance-level label. The cost function is iteratively minimized until some stopping criterion is reached.  Full details of the $e$FUMI algorithm can be found in the paper by Jiao and Zare \cite{jiao2015functions}.

\section{PROPOSED METHOD}
Given the $e$FUMI algorithm outlined in the previous section, our goal in this paper is to estimate the \emph{influence} of each pixel or an image region on the result of the $e$FUMI algorithm.  Specifically, our goal is to estimate which pixels would cause the largest change in the target spectral signature estimated by $e$FUMI if their labels were switched.  By identifying which pixels would cause the largest change in the $e$FUMI algorithm, then one could guide an analyst to focus on (re-)labeling only those pixels that would have a significant influence on the $e$FUMI result. Hopefully, this would result saving analyst time and effort during analysis. 

%The $e$FUMI-estimated target signature varies as the bag-level label changes for each training point. Therefore, \emph{influence}, defined as the L2 norm between the $e$FUMI-estimated target signature before and after label change, was proposed to measure the 
%The alteration of existing labels for $e$FUMI results in different target signature estimation. Therefore, in order to evaluate the degree of impact on $e$FUMI estimation caused by the label change, a metric called \emph{influence} is defined as the L2 norm between the $e$FUMI-estimated target signature before and after label change as in shown in (3). Generally, a larger influence value indicates the label change of this or these points have a larger impact for $e$FUMI target signature estimation, so called more influential (influential points).  
%Influence intuitively represents how much hyperspectral unmixing results changes, after relabeling a point or superpixel followed by rerunning $e$FUMI, compared with the case without relabeling, especially how much the target endmember signature changes. 

Given the goal of determining which pixels cause the largest change in the estimate target spectral signature, we define influence of a pixel as:
\begin{equation}
{I}_{i}={\lVert \mathbf{e}_{t_{i}}-\mathbf{e}_{t} \rVert}^{2}
\label{eqn:Influence}
\end{equation}
where $\mathbf{e}_{t}$ is the estimated target endmember without changing any label values (i.e., the target estimation result using an initial set of bag-level labels) and $\mathbf{e}_{t_{i}}$ is the estimated target endmember after changing the label of point $\mathbf{x}_{i}$ (i.e., switching the label to be ``negative'' to ``positive'' or vice versa.  One could certainly determine influence of each pixel by laboriously switching the label of each pixel in series, re-running $e$FUMI for the modified label set, and computing the resulting change in the estimated target signature.  However, this would be extremely time consuming and, thus, infeasible for the development of an interactive method for hyperspectral unmixing analysis.  

Therefore, this article proposes two fast methods to approximate the relative influence of a single point or image region (e.g., a superpixel). Both of the proposed methods extremely reduce the time needed to estimate relative influence as compared with original relabeling-rerunning process for each point/superpixel. Instead, only running $e$FUMI once is necessary. After running $e$FUMI with initial user-selected labels, the first proposed method estimates relative influence by appling the fully-constrained least-squares \cite {heinz2001fully} unmixing algorithm to all data points using the  endmembers esitmated from the initial run.  Then, we propose to use the target proportion value estimated for each point as a surrogate for the influence measure.  Namely, for each point $\mathbf{x}_{i}$, we find:
\begin{eqnarray}
& \mathbf{p}_{i}=\argmin_{\mathbf{p}_{i}} {\lVert {\mathbf{x}_{i}}-\mathbf{E}{\mathbf{p}_{i}} \rVert}_2^2\\
& s.t. \quad
\begin{cases}
{p}_{ij}\geq0 \quad \forall i,j\\
\sum_{j=1}^{M+1} p_{ij} \quad \forall i,j
\end{cases}
\label{eqn:unmix}
\end{eqnarray}
where $\mathbf{E}$ is a matrix containing the estimated endmembers, $\mathbf{p}_{i}$ is the proportion vector for point $\mathbf{x}_{i}$, $p_{ij}$ is the proportion value of endmember $j$, and $M$ is the number of non-target endmembers. The surrogate influence measure is set to $p_{it}$, the proportion value estimated for the target endmember for the $i^{th}$ data point. This method identifies the data points with large target portion.  Thus, changing the label of these data points would have a large influence of the resulting estimated target endmember. 

For the second proposed method, given the estimated endmembers and proportions for each point, the residual error is computed:
\begin{equation}
r_{i}={\lVert {\mathbf{x}_{i}}-\mathbf{E}{\mathbf{p}_{i}} \rVert}_2^{2}
\end{equation}
where $\mathbf{E}$ is the estimated endmembers, $\mathbf{p}_{i}$ is the estimated proportion for point $\mathbf{x}_{i}$.
This method identifies the data  points that are not well represented by the estimated endmembers and, thus, lie outside the convex hull defined by the endmembers. % Since $e$FUMI is robust to the error caused by mislabeling a few true target points \cite {jiao2015functions}, the influence of single target point are not considered in this article. However, \cite {jiao2015functions} indicates that when the number of mislabeled target points reach some threshold, the $e$FUMI estimation results do have a considerable change. Therefore, in the superpixel experiment part, we also flip the labels of target points since a superpixel may contains many target points.

%The definition of influence. \\ The definition of target proportion by unmixing. \\ The definition of residual error. \\ Some explanation about the relationship between influence and proposed methods.

\section{EXPERIMENTAL RESULTS}

Experimental evaluations have been carried out to investigate the performance of the proposed methods.  

\subsection{Datasets}
The proposed instance influence estimation methods were tested on two hyperspectral datasets.

\subsubsection{MUUFL Gulfport Data Set} The first hyperspectral data set used in experiments in this paper was acquired by the CASI-1500 hyperspectral imager over the campus of the University of Southern Mississippi-Gulfpark in Long Beach, Mississippi in November 2010. The scene contains $325 \times 337$ pixels and consists of 72 bands covering a wavelength range of 375 to 1050 nm with a 10 nm bandwidth. The spatial resolution is 1 m. Four different types of colored fabric cloths were placed throughout the scene as targets.  The target colors (used to identify the different target types) are Brown, Dark Green, Faux Vineyard Green, and Pea Green. Each type of target varied in size from $0.5 \times 0.5$ $m^{2}$, $1 \times 1$ $m^{2}$, to $3 \times 3$ $m^{2}$. A 5-by-5 halo around the center of each target was considered as the target ``bag'' since the accuracy of the GPS device used to collect the ground truth was 5 m. \cite {gader2013muufl} 

\subsubsection{Pavia University Data Set} The second data set was collected by the Reflective Optics System Imaging Spectrometer (ROSIS) around the Engineering School at the University of Pavia, Italy in July 2002. The flight was operated by the Deutschen Zentrum for Luftund Raumfahrt (DLR, the German Aerospace Agency) in the framework of the HySens project, managed and sponsored by the European Union.\cite {dopido2013semisupervised} The ROSIS generates 115 bands cover 430 to 860 nm with 4nm bandwidth. The scene contain $340 \times 610$ pixels and consists of 103 bands with 12 noisiest bands removed. The spatial resolution is 1.3 m. We obtained this dataset from the website of Computational Intelligence Group at the Basque University (UPV/EHU). \cite {pavia}

The target of interest for this data set was selected as a region of sidewalk (found around the blue painted metal sheets). The label of target bag and non-target bags were chosen to ensure that the target bag contains the target material and non-target bags contained a variety of non-target materials such as painted metal sheets, vegetation, red roof, bare soil and shadow. 

\begin{figure}
\begin{center}
\hfill
\subfigure[]{\includegraphics[height=5cm]{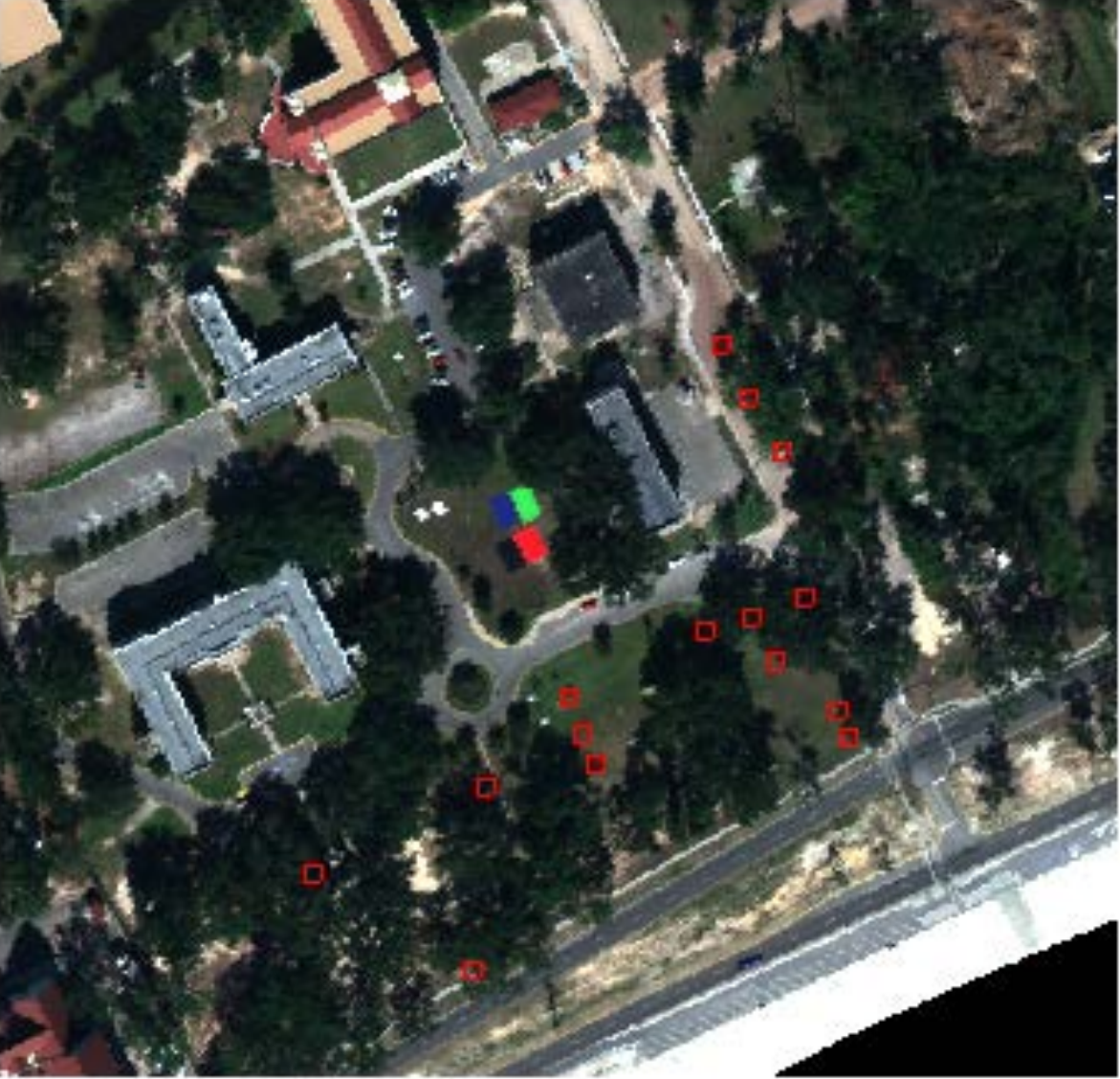}}
\hfill
\subfigure[]{\includegraphics[height=5cm]{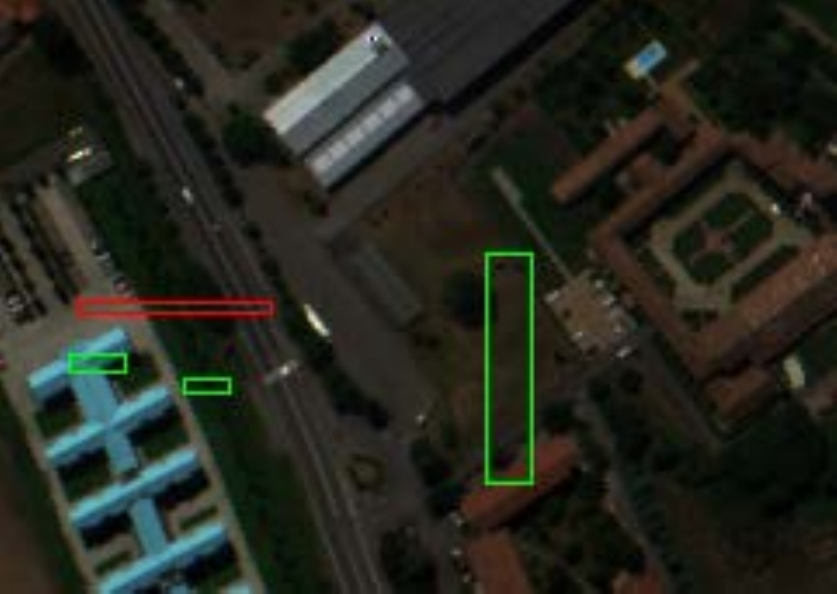}}
\hfill
\caption{Labeled ground truth on (a) Gulfport. 5-by-5 red regions are the target bags (Brown) and the rest are non-target bags; (b) Pavia University. Red region is the target bag (Sidewalk) and green regions are the non-target bags}
\label{fig:label}
\end{center}
\end{figure}

\subsection{Single Point Experiment}

The first experiment investigated the correlation between the influence of a single data point and the proposed metrics. 
 The $e$FUMI was first run given an initial set of labels (as shown in Figure \ref{fig:label}).  When running $e$FUMI, endmembers were initialized using the VCA algorithm \cite {nascimento2005vertex} and all proportion values were initialized as $\frac{1}{M+1}$ where $M$ is the number of background endmembers and $M+1$ is the total number of endmembers (given only one target endmember).  After this initial run, both of the proposed methods to approximate relative influence were computed.  Thus, the target proportion value and the residual error were computed for every data point.  Then, the labels of 1000 non-target data points were flipped in series. For each data point, $e$FUMI was run again given the new label set and the resulting influence value was computed.  After the influence value was computed for a data point, the label of that point was restored and the next label was flipped to repeat the same process such that the influence values for all 1000 points were calculated. This process is outlined in Alg. \ref{alg:singlepoint}. 

\begin{algorithm}

\caption{Single Point Experiment}
\algsetup{indent=4em}
\begin{algorithmic}[1]  
\STATE Initialize $\mathbf{E}_{init}$ via VCA, $\mathbf{P}_{init}$ as $\frac{1}{M+1}$
\STATE $\mathbf{E}_{0},\mathbf{P}_{0} \gets $ $e$FUMI($L$, $\mathbf{X}$, $\mathbf{E}_{init}$, $\mathbf{P}_{init}$) (where $L$ is the initial label set and $\mathbf{X}$ is the data)
\STATE Compute residual error for each data point: $\mathbf{r} \gets \left\| \mathbf{X} - \mathbf{E}_{0}\mathbf{P}_0 \right\|_2^2$
\STATE Obtain target proportion for each data point, $\mathbf{p}_t$ via unmixing using \eqref{eqn:unmix}
\FOR{i=1:NumDataPoints} 
\STATE Flip the label $\mathbf{l}_{i}$ of point $\mathbf{x}_{i}$ to obtain updated label set, $L^i$:  $\mathbf{l}_{i} \gets  \mathbf{l}_{i} -1$ 
\STATE $\mathbf{E}_{i},\mathbf{P}_{i} \gets $ $e$FUMI($L^i$, $\mathbf{X}$, $\mathbf{E}_{0},\mathbf{P}_{0}$)
\STATE Compute Influence: $\mathbf{I}_{i}$ using \eqref{eqn:Influence}
\STATE Restore the label $\mathbf{l}_{i}$:  $\mathbf{l}_{i} \gets  \mathbf{l}_{i} -1$ 
\ENDFOR

   \RETURN \\
\end{algorithmic} 
\label{alg:singlepoint}
\end{algorithm}

\begin{figure}
\begin{center}
\hfill
\subfigure[]{\includegraphics[height=3cm]{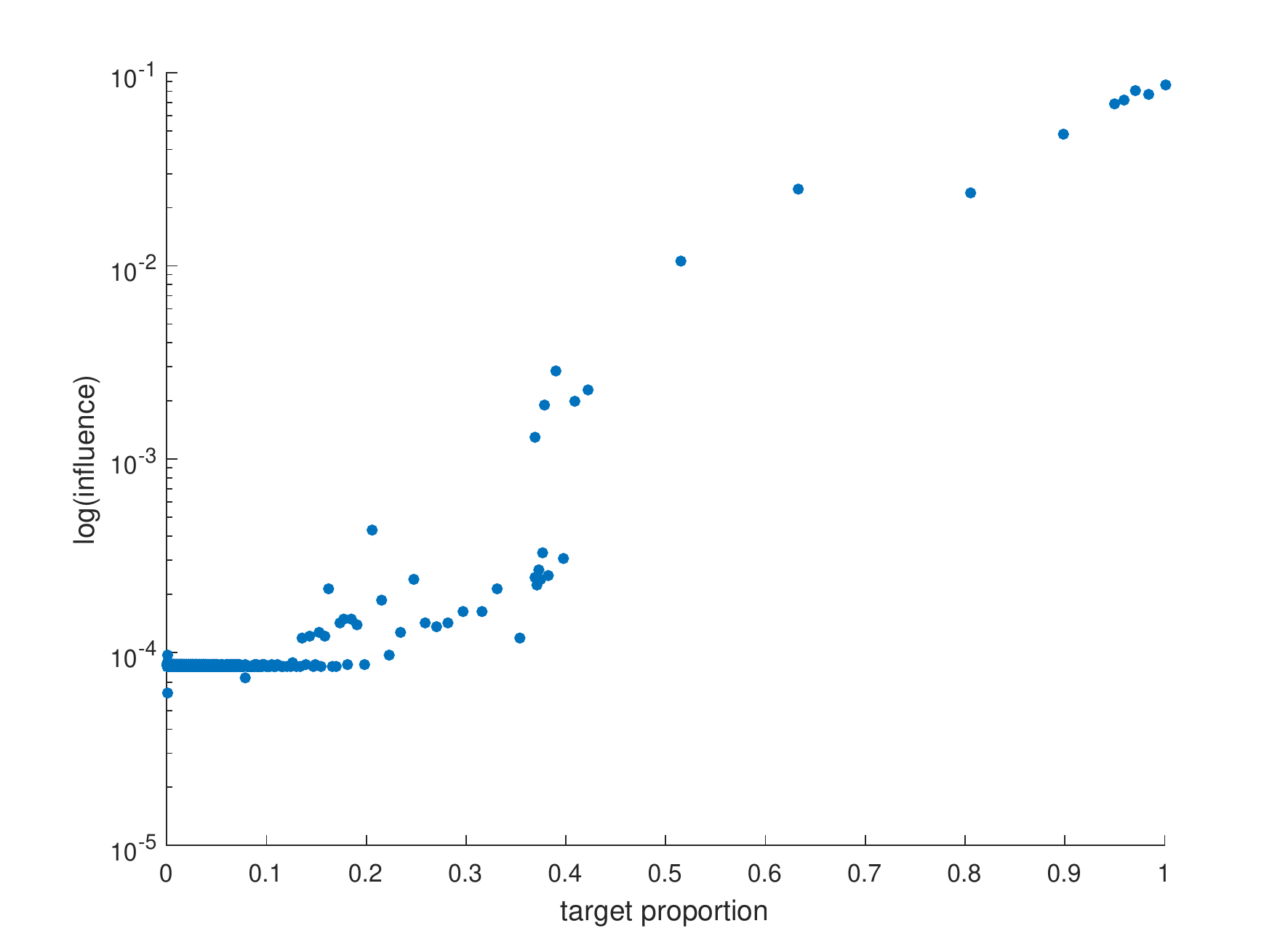}}
\hfill
\subfigure[]{\includegraphics[height=3cm]{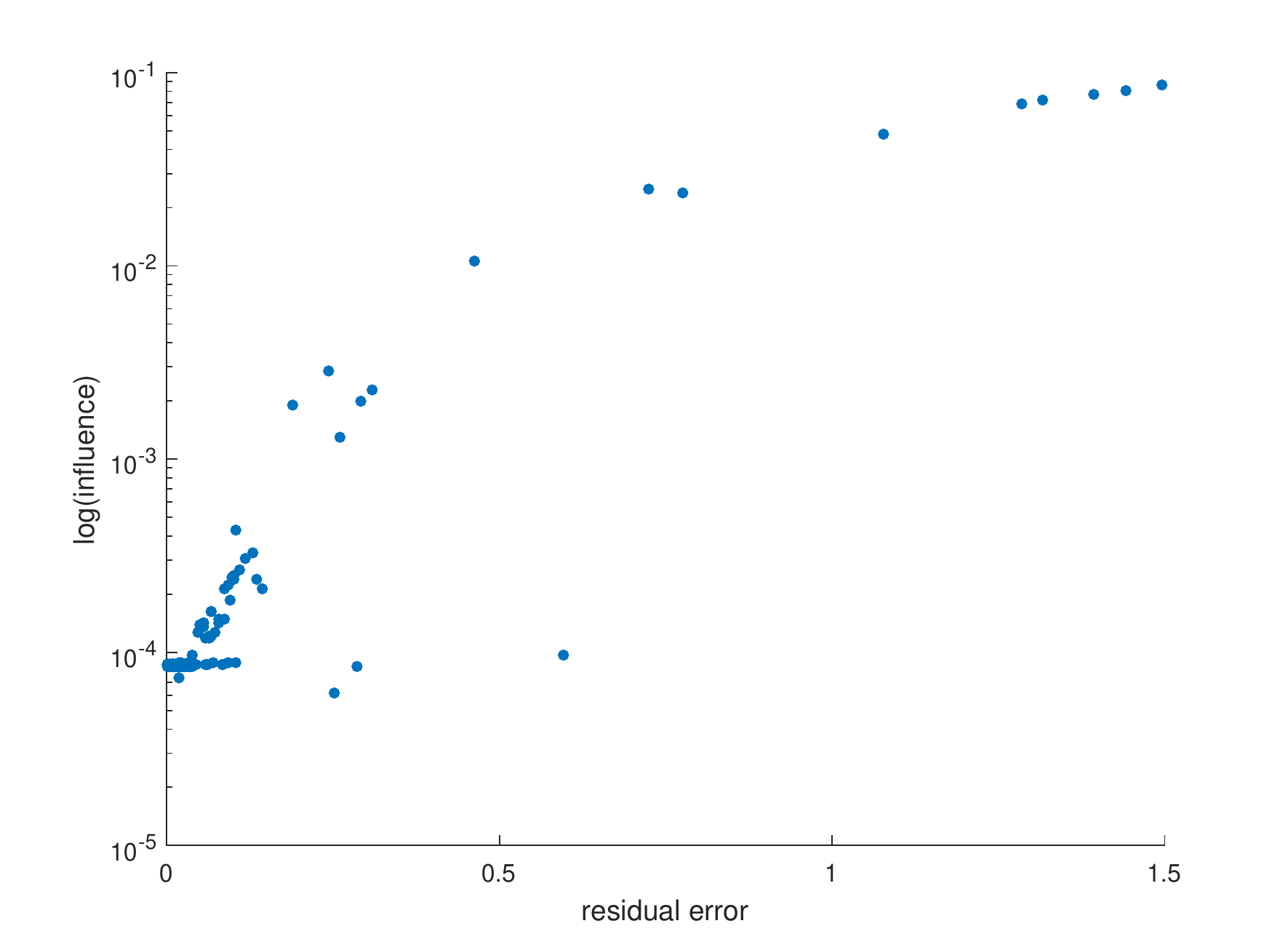}}
\hfill
\subfigure[]{\includegraphics[height=3cm]{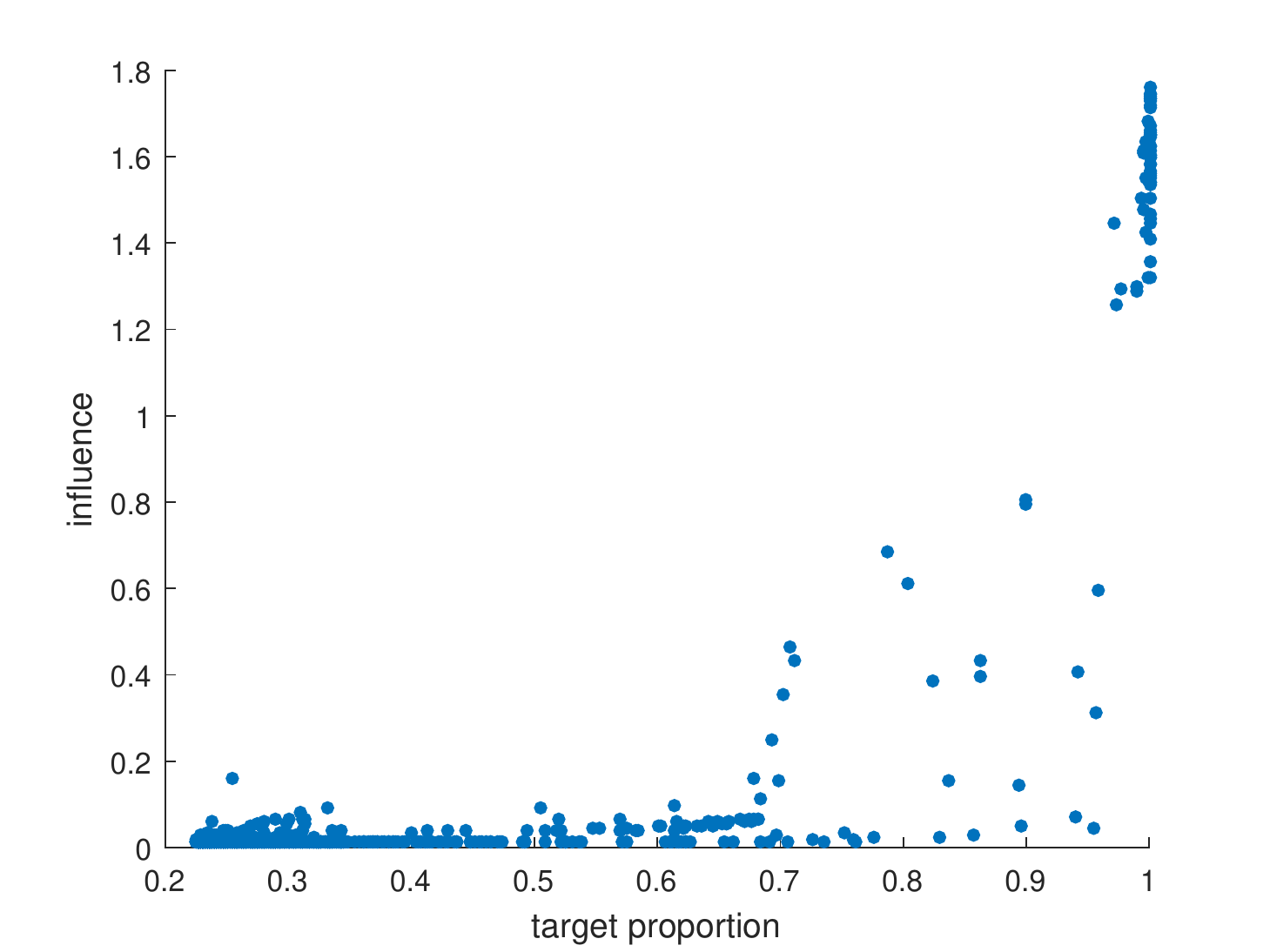}}
\hfill
\subfigure[]{\includegraphics[height=3cm]{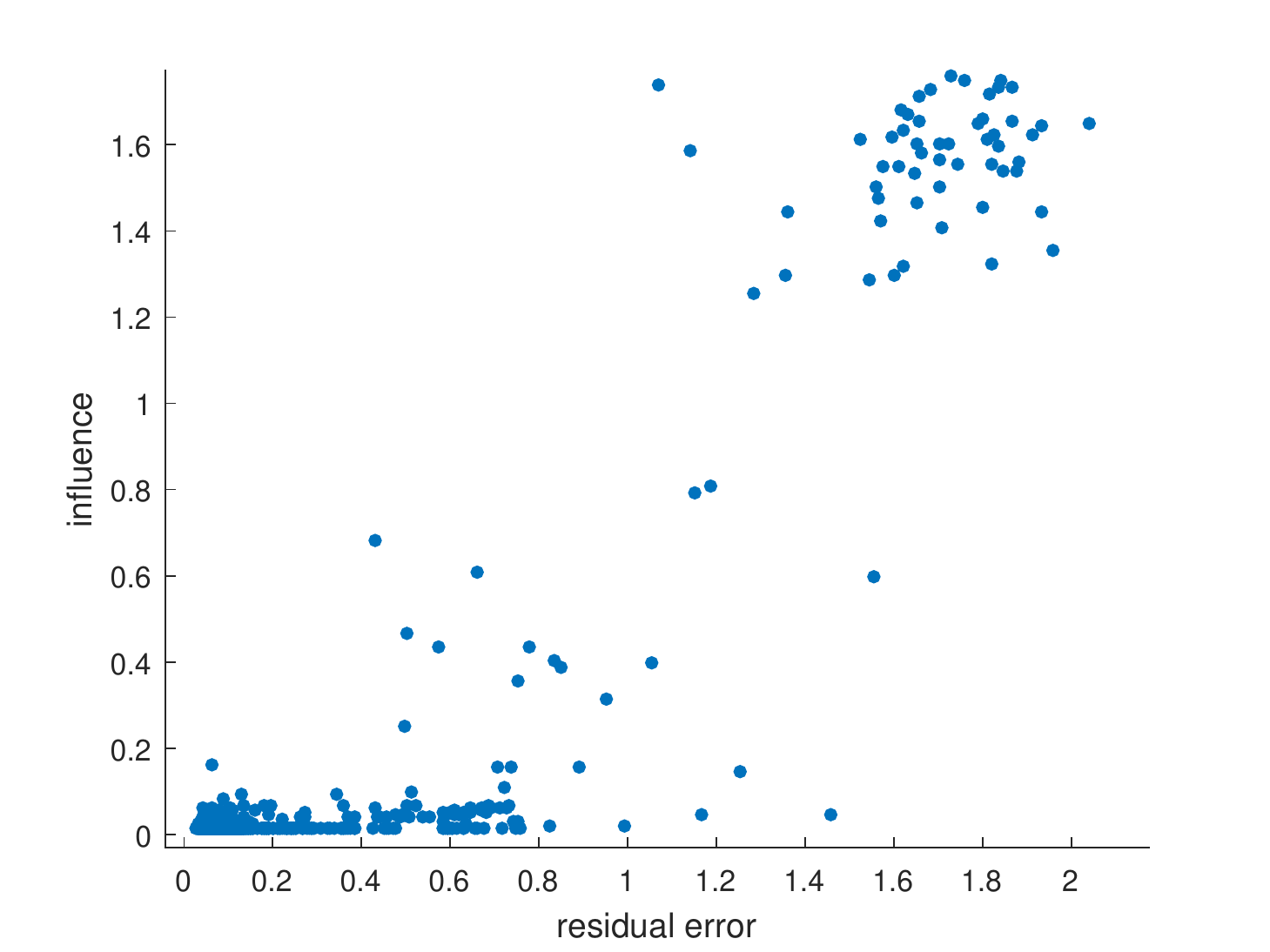}}
\hfill
\caption{Influence caused by selecting 1000 points randomly(a,b) or with high target proportion(c,d) from Gulfport and changing labels, compared with their corresponding (a,c) target proportion, (b,d) residual error}
\label{fig:single_influence_Gulfport}
\end{center}
\end{figure}

  Figure \ref{fig:single_influence_Gulfport} (a) and (b) display the scatter plots of log of the influence value versus the target proportion and residual error metrics,
respectively, for 1000 randomly selected points. The Spearman\rq s rank correlation coefficients for both methods are $\rho_{p_{t}}=0.3875$ and $\rho_{re}=0.1005$.  Figure \ref{fig:single_influence_Gulfport} (c) and (d) display the scatter plots of log of the influence value versus the target proportion and residual error metrics,
respectively, for the 1000 data points with the largest target proportion value. The Spearman\rq s rank correlation coefficients for both methods are $\rho_{p_{t}}=0.6434$ and $\rho_{re}=0.5674$. As can be seen in the scatter plots, points with large influence values tend to also have large target proportion and large residual error values.

\begin{figure}
\begin{center}
\hfill
\subfigure[]{\includegraphics[height=3cm]{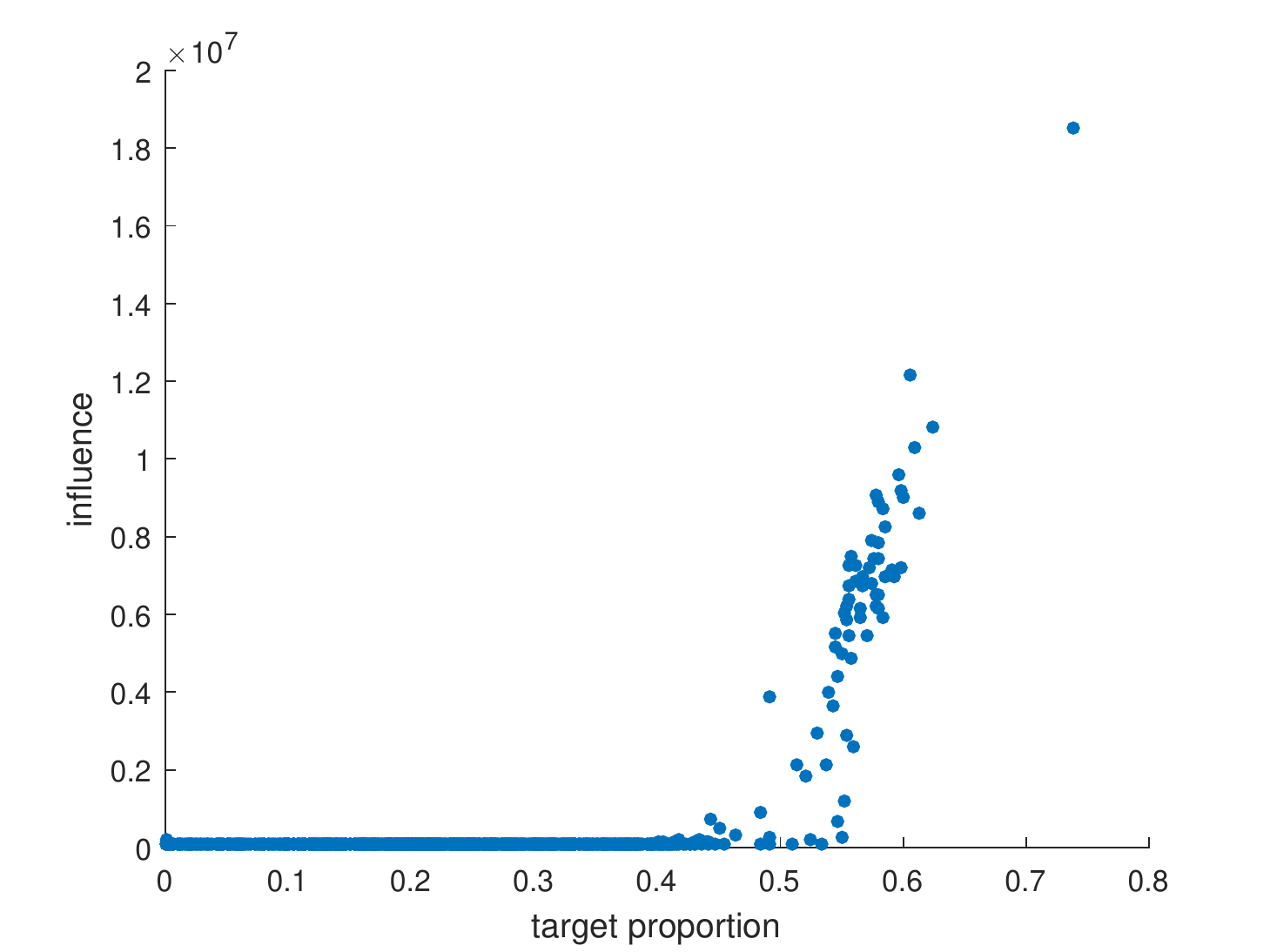}}
\hfill
\subfigure[]{\includegraphics[height=3cm]{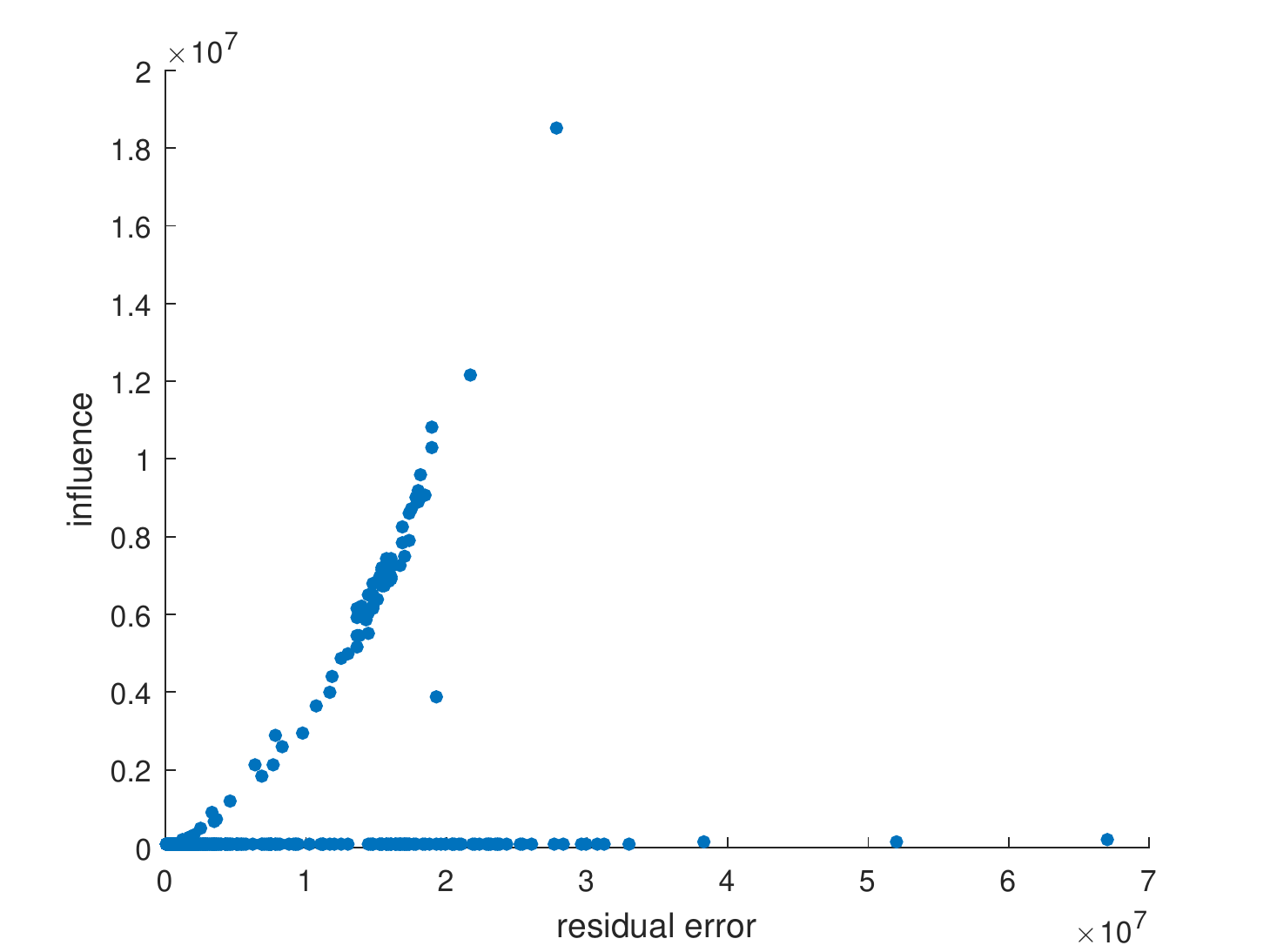}}
\hfill
\caption{Influence caused by selecting all data points in the non-target bags with the large target proportion value from the Pavia data set and changing labels, compared with their corresponding (a) target proportion, (b) residual error}
\label{fig:single_influence_Pavia}
\end{center}
\end{figure}

The proposed methods are also shown to be effective on the Pavia University dataset. As is shown in Figure \ref{fig:label} (b), a sub-image of Pavia University was chosen as the test dataset. The red rectangular region was marked as the target bag and green rectangular regions were identified as the non-target bags. In this example,  the target material was ``sidewalk''.  The labels of all non-target data points were flipped sequentially, following the procedure described above.  As shown in Figure \ref{fig:single_influence_Pavia} (a), there is a strong relationship between high influence value and high target proportion on this data set. Figure \ref{fig:single_influence_Pavia} (b) displays the relationship between influence value and residual error. The data points that have a target proportion above 0.4 in Figure \ref{fig:single_influence_Pavia} (a) were identified as pixels from red roof regions in the non-target bags. Further investigation shows that these red roof materials have a similar spectral signature to the sidewalk in the area. In this way, act as a ``confuser'' during target characterization and, thus, they are influnetial data points in the $e$FUMI algorithm. In this example, target portion was seen to be more effective than residual error in predicting high influence values (as can bee seen with the divergence of influence valus in the scatter plot among points with large residual error).  

\subsection{Mislabeling and Recovery Experiment}

In this experiment, we investigate the improvement in target endmember estimation by correcting the labels of groups of mislabeled data points.  Specifically, we compare the improvement of selecting points to relabel using the proposed methods versus random selection. 

The experiment was structured as follows.  First, $e$FUMI was run with a fixed set of correct initial labels. The resulting endmembers and proportion values were labeled as $E_{true}$ and $P_{true}$.   Then, 0.5\% of the labels (500 non-target points) were randomly selected and changed to incorrect label values.  $e$FUMI was run using these incorrect labels and the results were named $E_{err}$ and $P_{err}$. Finally, 20\% data points were selected using one of three different strategies: (1) selected 20\% of the data points randomly, (2) selected 20\% of the  data points with largest target proportion, and (3) selected 20\% of the data points with the largest residual error. Among the 20\% of selected pixels, their labels were corrected if incorrect.  These updated labels were then used to run $e$FUMI. These final set of estimated endmembers were named $E_{rand}$, $E_{p_{t}}$, and $E_{re}$, respectively, depending on the sample selection strategy. This process is outlined in Alg. \ref{alg:mislabeling}. 

\begin{algorithm}

\caption{Mislabeling and Recovery Experiment}
\algsetup{indent=4em}
\begin{algorithmic}[1]  
\STATE Initialize $\mathbf{E}_{init}$ via VCA, $\mathbf{P}_{init}$ as $\frac{1}{M+1}$
\STATE $\mathbf{E}_{true},\mathbf{P}_{true} \gets $ $e$FUMI($L$, $\mathbf{X}$, $\mathbf{E}_{init}$, $\mathbf{P}_{init}$) (where $L$ is the initial label set and $\mathbf{X}$ is the data)
\STATE Compute residual error for each data point: $\mathbf{r} \gets \left\| \mathbf{X} - \mathbf{E}_{0}\mathbf{P}_0 \right\|_2^2$
\STATE Obtain target proportion for each data point, $\mathbf{p}_t$ via unmixing using \eqref{eqn:unmix}
\STATE Randomly select and flip $\alpha\%$ labels obtaining $L^{err}$
\STATE $\mathbf{E}_{err},\mathbf{P}_{err} \gets $ $e$FUMI($L^{err}$, $\mathbf{X}$, $\mathbf{E}_{init}$, $\mathbf{P}_{init}$) (where $L$ is the updated label set and $\mathbf{X}$ is the data)
\STATE Initialize $\mathbf{E}_0$ via VCA, $\mathbf{P}_0$ as $\frac{1}{M+1}$
\FOR{i=1 to 3}
\IF{i=1}
\STATE Randomly select $\beta\%$ labels and correct if wrong to obtain $L^{rand}$
\STATE $\mathbf{E}_{rand},\mathbf{P}_{rand} \gets $ $e$FUMI($L^{rand}$, $\mathbf{X}$,  $\mathbf{E}_0$, $\mathbf{P}_0$)
\STATE Compute $DoI_{rand}$ using \eqref{eqn:DoI}
\ENDIF
\IF{i=2}
\STATE Select points with highest $\beta\%$ target proportions and correct their labels if wrong to obtain $L^{p_t}$
\STATE $\mathbf{E}_{p_{t}},\mathbf{P}_{p_{t}} \gets $ $e$FUMI($L^{p_t}$, $\mathbf{X}$,  $\mathbf{E}_0$, $\mathbf{P}_0$)
\STATE Compute $DoI_{p_{t}}$ using \eqref{eqn:DoI}
\ENDIF
\IF{i=3}
\STATE Select points with highest $\beta\%$ residual errors and correct their labels if wrong to obtain $L^{re}$
\STATE $\mathbf{E}_{re},\mathbf{P}_{re} \gets $ $e$FUMI($L^{re}$, $\mathbf{X}$,  $\mathbf{E}_0$, $\mathbf{P}_0$) 
\STATE Compute $DoI_{re}$ using \eqref{eqn:DoI}
\ENDIF
\ENDFOR
   \RETURN \\
\end{algorithmic} 
\label{alg:mislabeling}
\end{algorithm}

\begin{figure}
\begin{center}
\hfill
\subfigure[Brown]{\includegraphics[height=3cm]{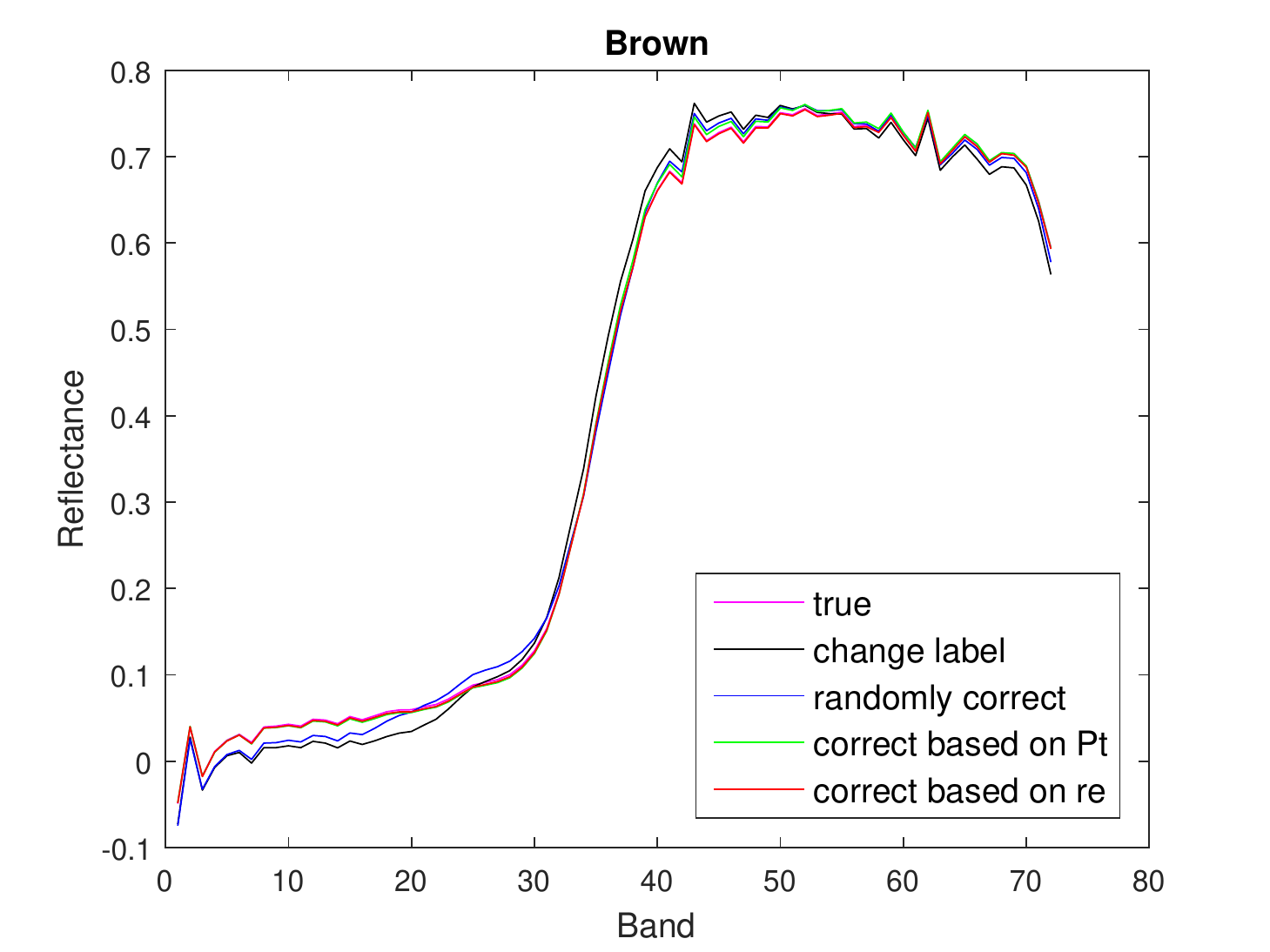}}
\hfill
\subfigure[Dark Green]{\includegraphics[height=3cm]{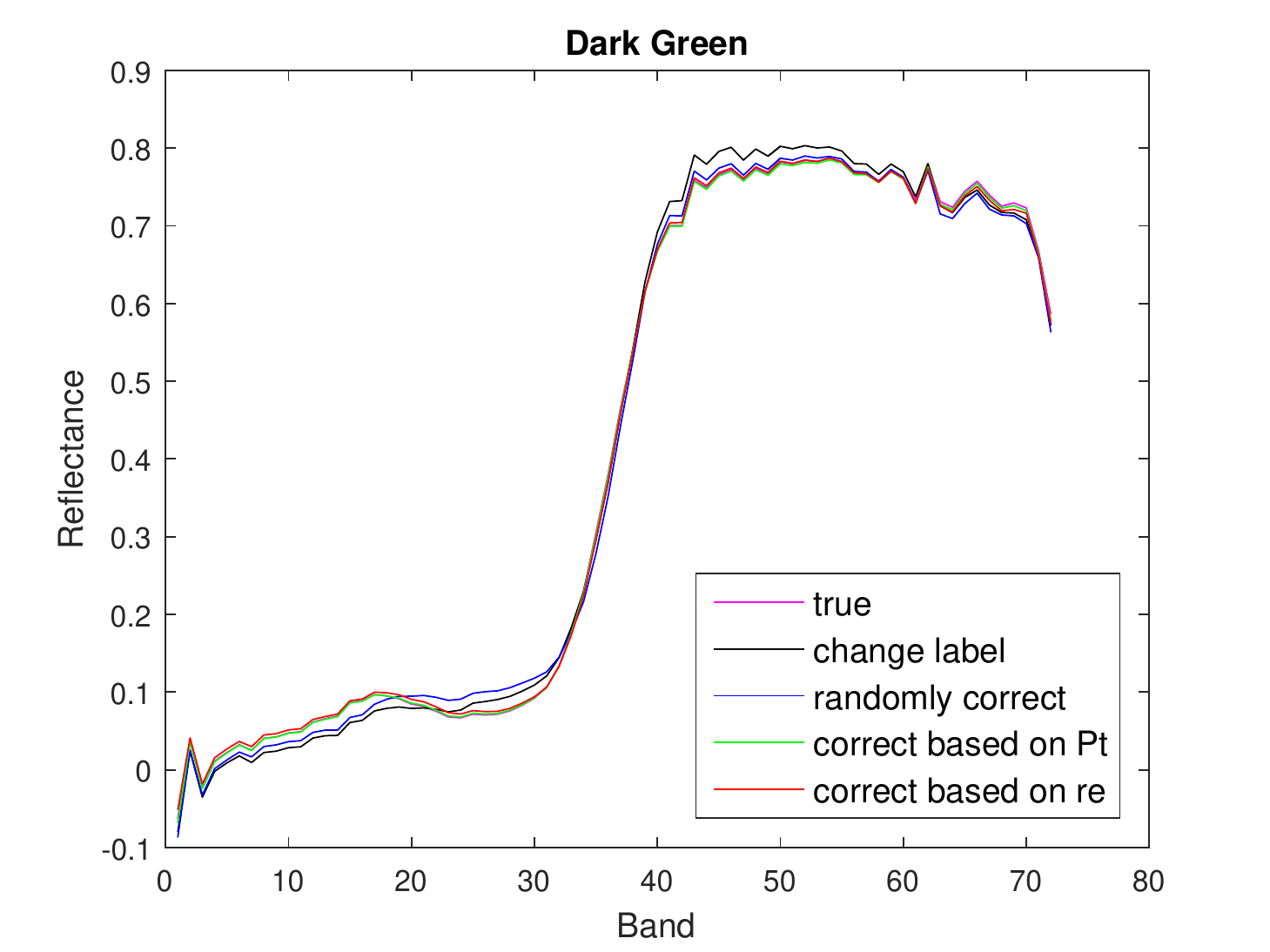}}
\hfill
\subfigure[Faux Vineyard Green]{\includegraphics[height=3cm]{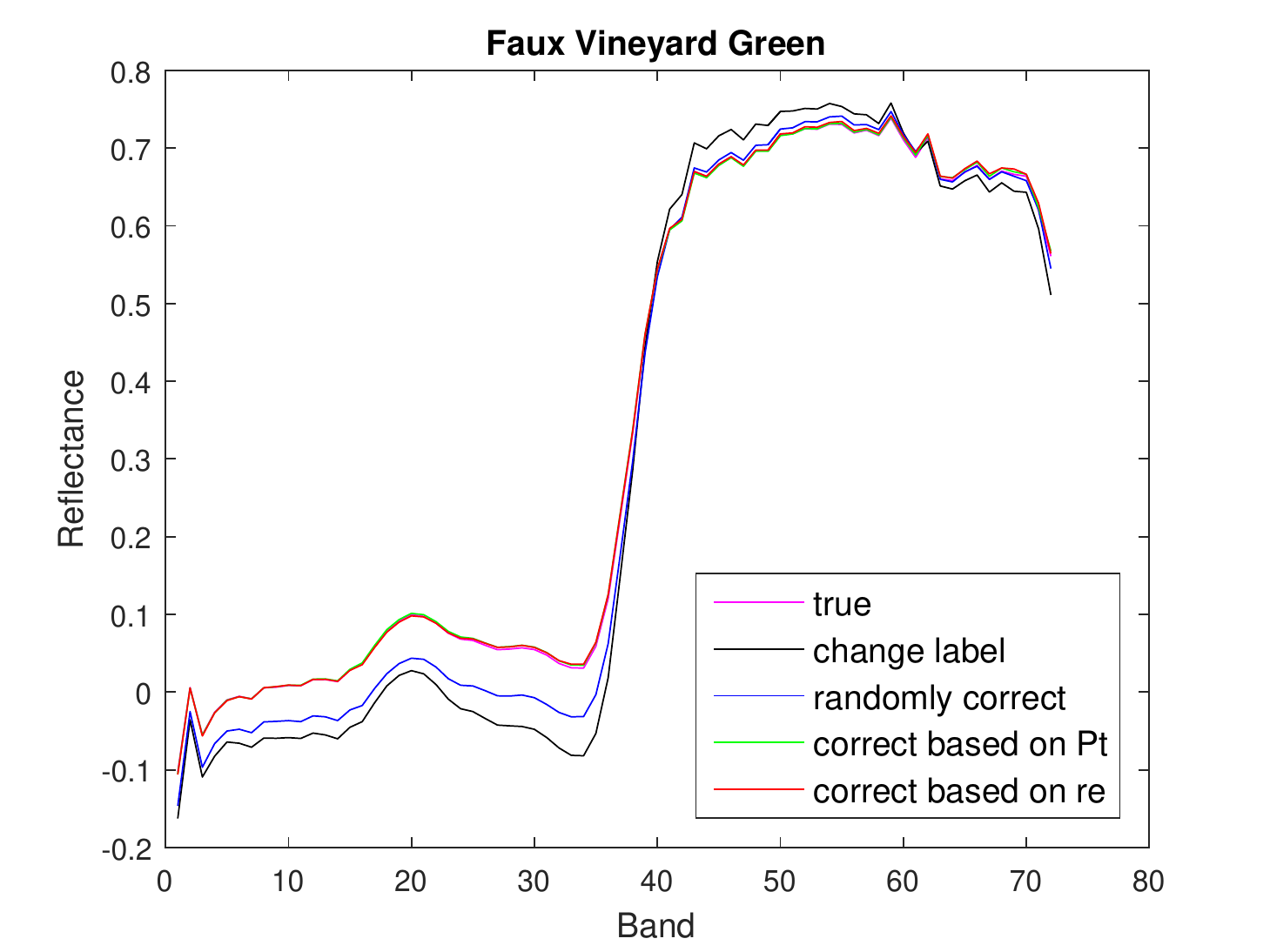}}
\hfill
\subfigure[Pea Green]{\includegraphics[height=3cm]{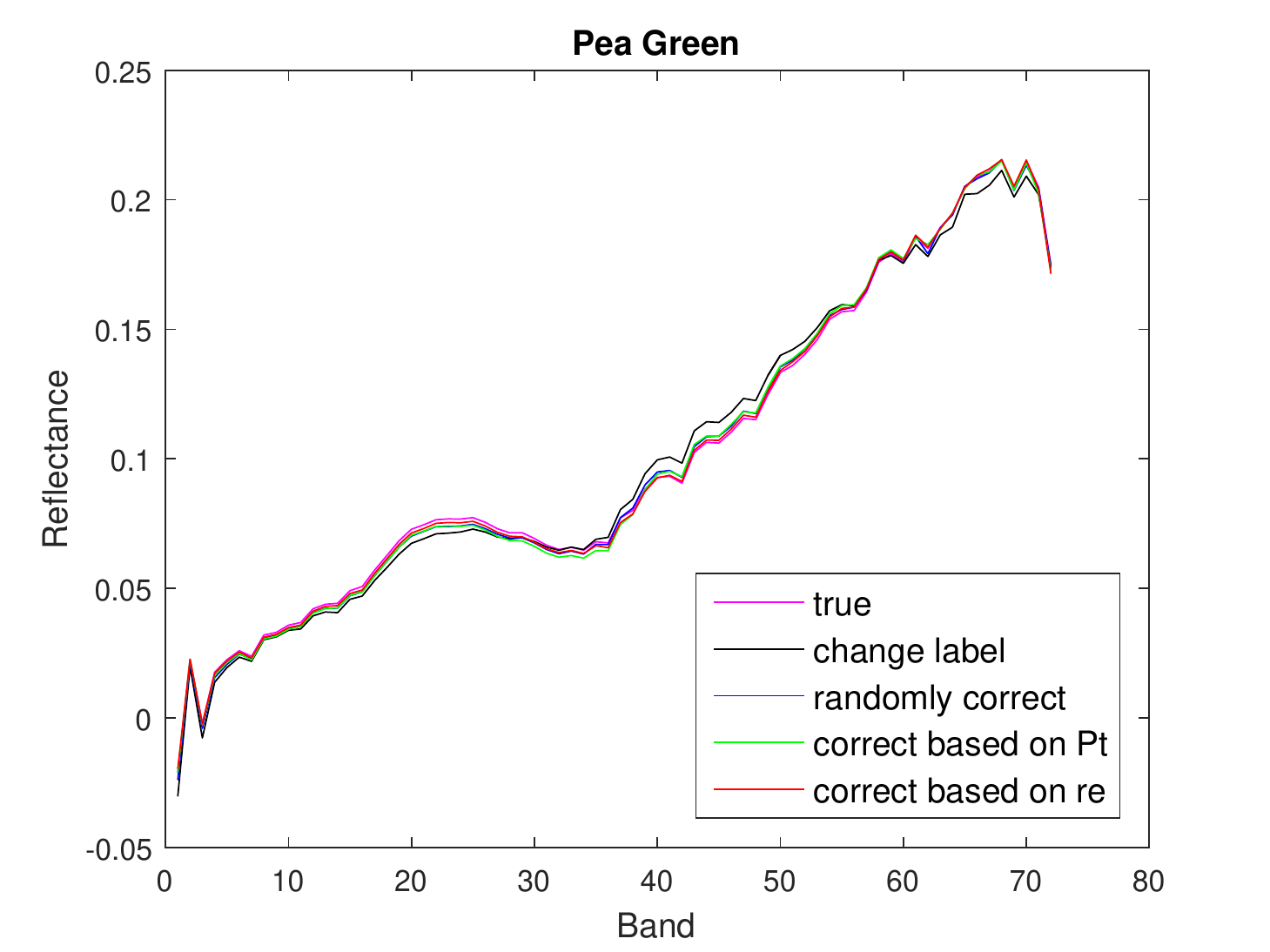}}
\hfill
\caption{Target endmembers on Gulfport estimated by label of (1) ground truth; (2) 0.5\% flipped; (3) correct 20\% randomly; (4) correct 20\% based on influential points of method 1; (5) correct 20\% based on influential points of method 2 }
\label{fig:mislabeling_Gulfport}
\end{center}
\end{figure}

 Degrees of improvement (DoI) for the target endmember spectra  for the three methods (random, target proportion and residual error) were calculated using the following:
\begin{equation}
DoI_{k}=\frac{{\lVert e_{t_{true}}-e_{t_{err}} \rVert}^{2}-{\lVert e_{t_{true}}-e_{t_{k}} \rVert}^{2}}{{\lVert e_{t_{true}}-e_{t_{err}} \rVert}^{2}}, k=rand, P_{t}, re,
\label{eqn:DoI}
\end{equation}
where $e_{t_{m}}$ is the target endmember for $e_{m}=true, err, rand, P_{t}, re$.  DoI for the three methods were computed as $DoI_{rand}=18.7\%$, $DoI_{P_{t}}=96.11\%$, $DoI_{re}=99.42\%$, respectively, considering the brown target in the MUUFL Gulfport data set. Figure \ref{fig:mislabeling_Gulfport} and the resulting DoIs show that the label correction obtained using the proposed methods could better recover the estimated target signature as opposed to random selection.

%\begin{figure}
%\begin{center}
%\includegraphics[height=6cm]{Figures/mis3correct30.eps}
%\caption{Degree of Improvement (DoI) for random and proposed methods with different targets on Gulfport}
%\end{center}
%\end{figure}

%Not only labeling Brown as the target, but alternatively labeling Dark Green, Faux vineyard green, Pea green as the target respectively to further demonstrate the effectiveness of proposed methods. As is shown in Figure 7, under the condition of mislabeling 3\% and correcting 30\%, both proposed methods achieve very high DoI compared with random correction. 

The mislabeling and recovery experiment was repeated on the Pavia Dataset. 5\% of the labels (58 non-target points) was chosen as the mislabeling ratio and 10\% was chosen as the correction ratio. The initial labels for $e$FUMI was the same one shown in Figure \ref{fig:label} (b). The DoI for three methods were found to be $DoI_{rand}=25.26\%$, $DoI_{P_{t}}=73.4\%$, and $DoI_{re}=28.06\%$. Similar to the single point experiment, it was found that the target proportion metric was more effective on this data set. 

% \begin{figure}
% \begin{center}
% \includegraphics[height=6cm]{Figures/Mislabel_pavia.eps}
% \caption{Target endmember on Pavia estimated by label of (1) ground truth; (2) 5\% flipped; (3) correct 10\% randomly; (4) correct 10\% based on influential points of method 1; (5) correct 10\% based on influential points of method 2}
% \label{fig:mislabeling_Pavia}
% \end{center}
% \end{figure}

\subsection{Superpixel Influence Experiment}

A superpixel is defined as a small, spatially continuous segment in an image. In this experiment, we investigate the influence of modifying labels of superpixels (as opposed to invidiual pixels) in a hyperspectral data set.  The influence for superpixel was investigated since each superpixel generally contains data points with similar spectral signatures (and, thus, the influence of the set of data points is likely to be similar) and because it is much easier and more intuitive to relabel superpixel regions instead of individual pixels. 

\begin{figure}
\begin{center}
\hfill
\subfigure[]{\includegraphics[height=5cm]{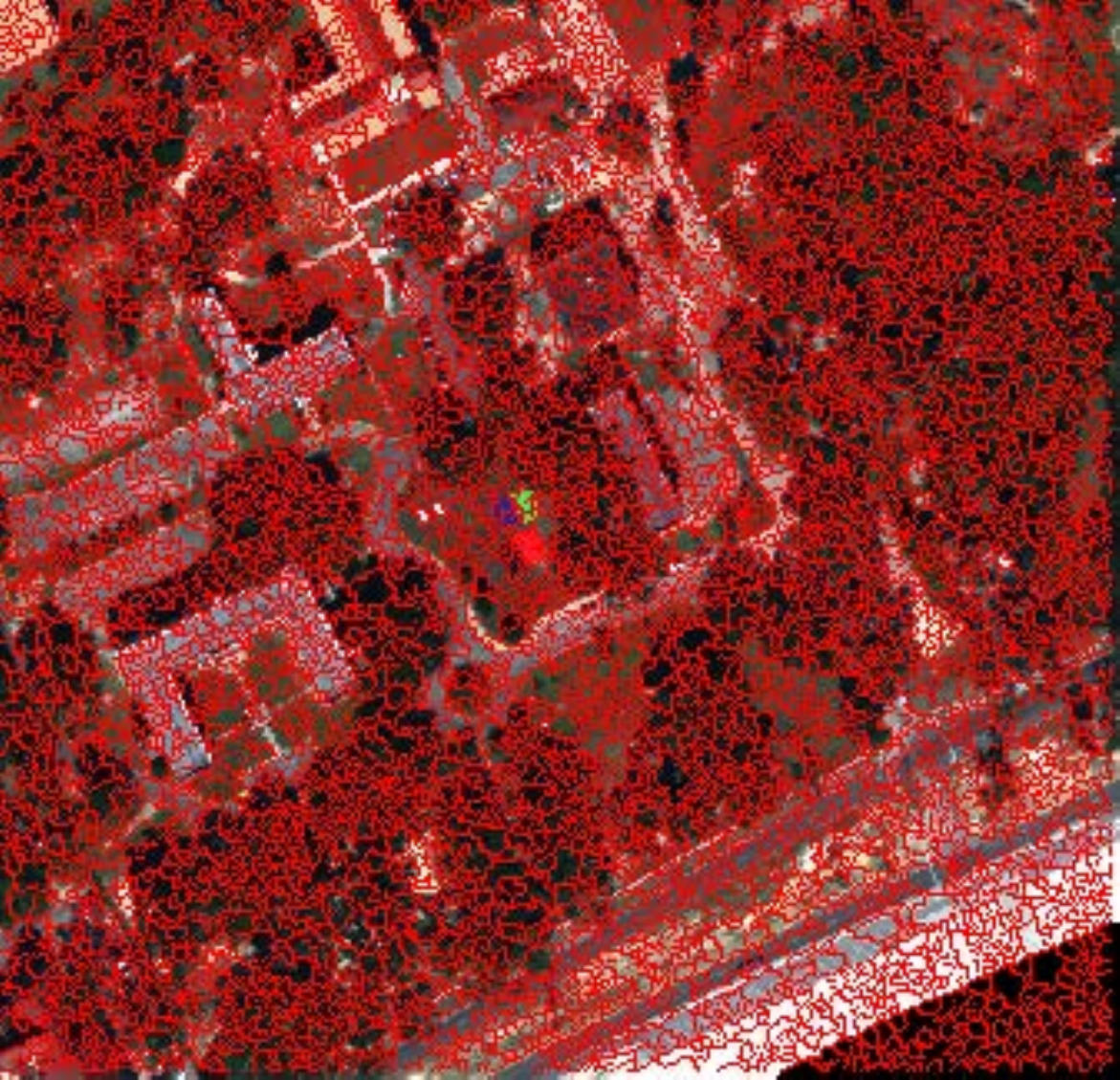}}
\hfill
\subfigure[]{\includegraphics[height=5cm]{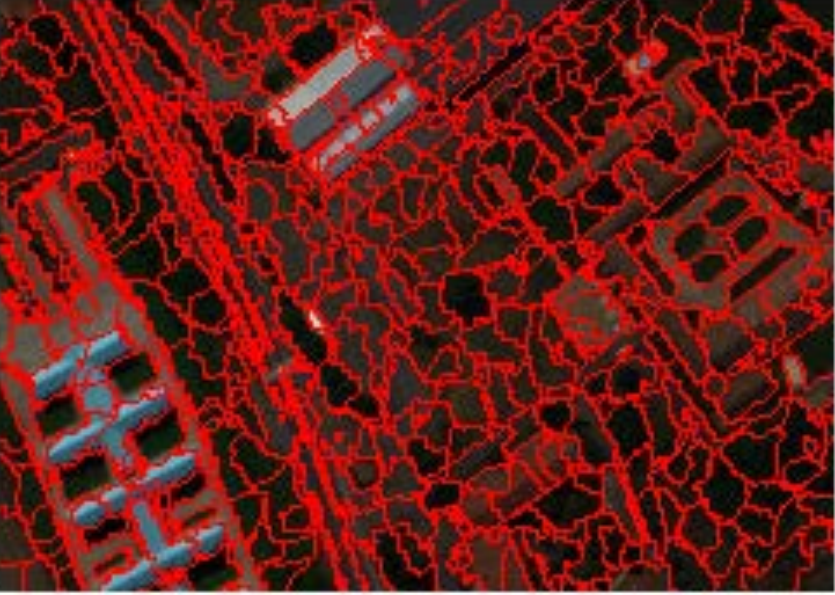}}
\caption{Influence of Superpixel for (a) Gulfport. 7419 segments were generated via Normalize Cut \cite {gillis2012hyperspectral}; (b) Pavia University. 744 segments were generated via Normalize Cut}
\label{fig:superpixel}
\end{center}
\end{figure}

\begin{figure}
\begin{center}
\hfill
\subfigure[]{\includegraphics[height=3cm]{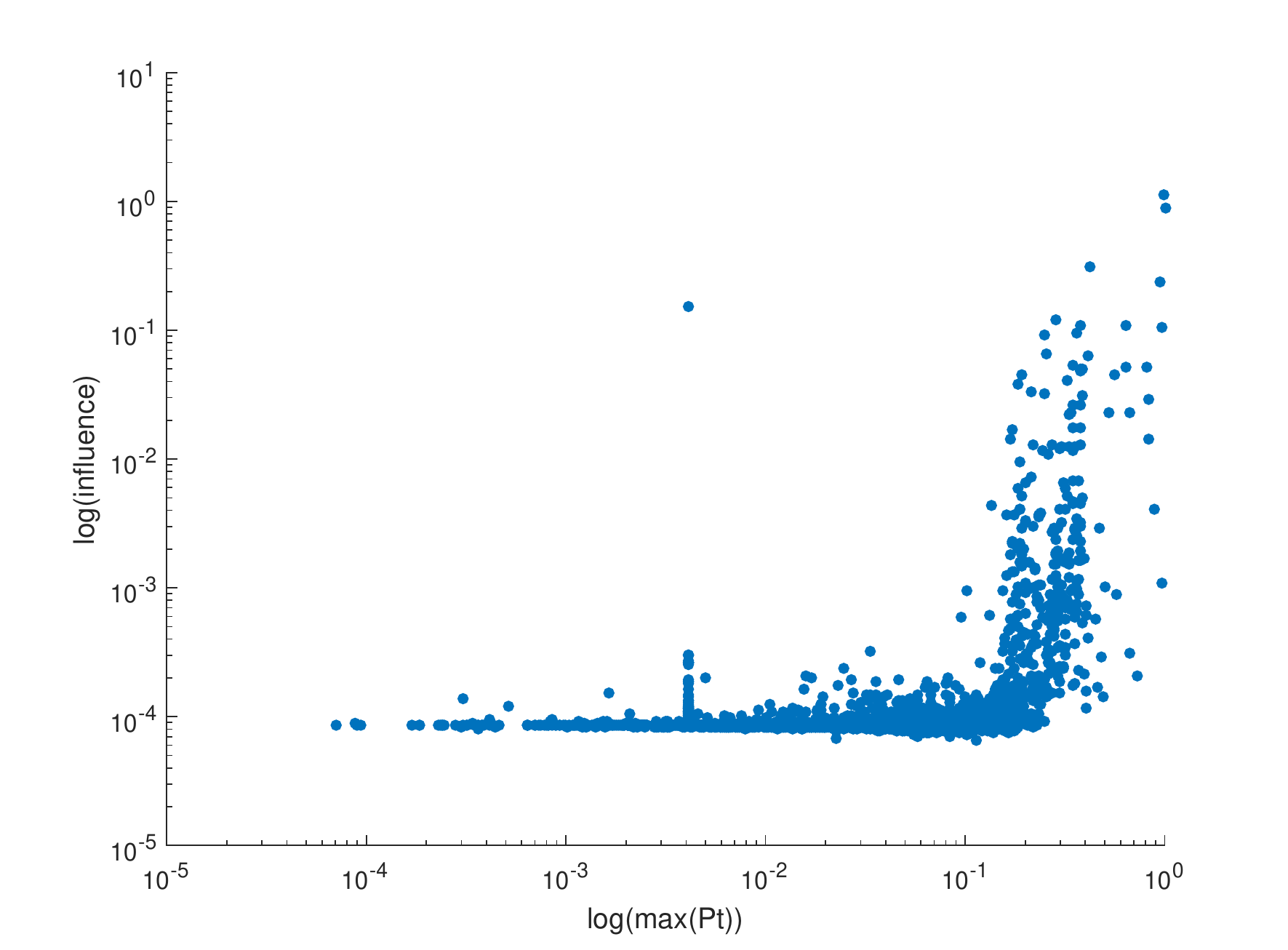}}
\hfill
\subfigure[]{\includegraphics[height=3cm]{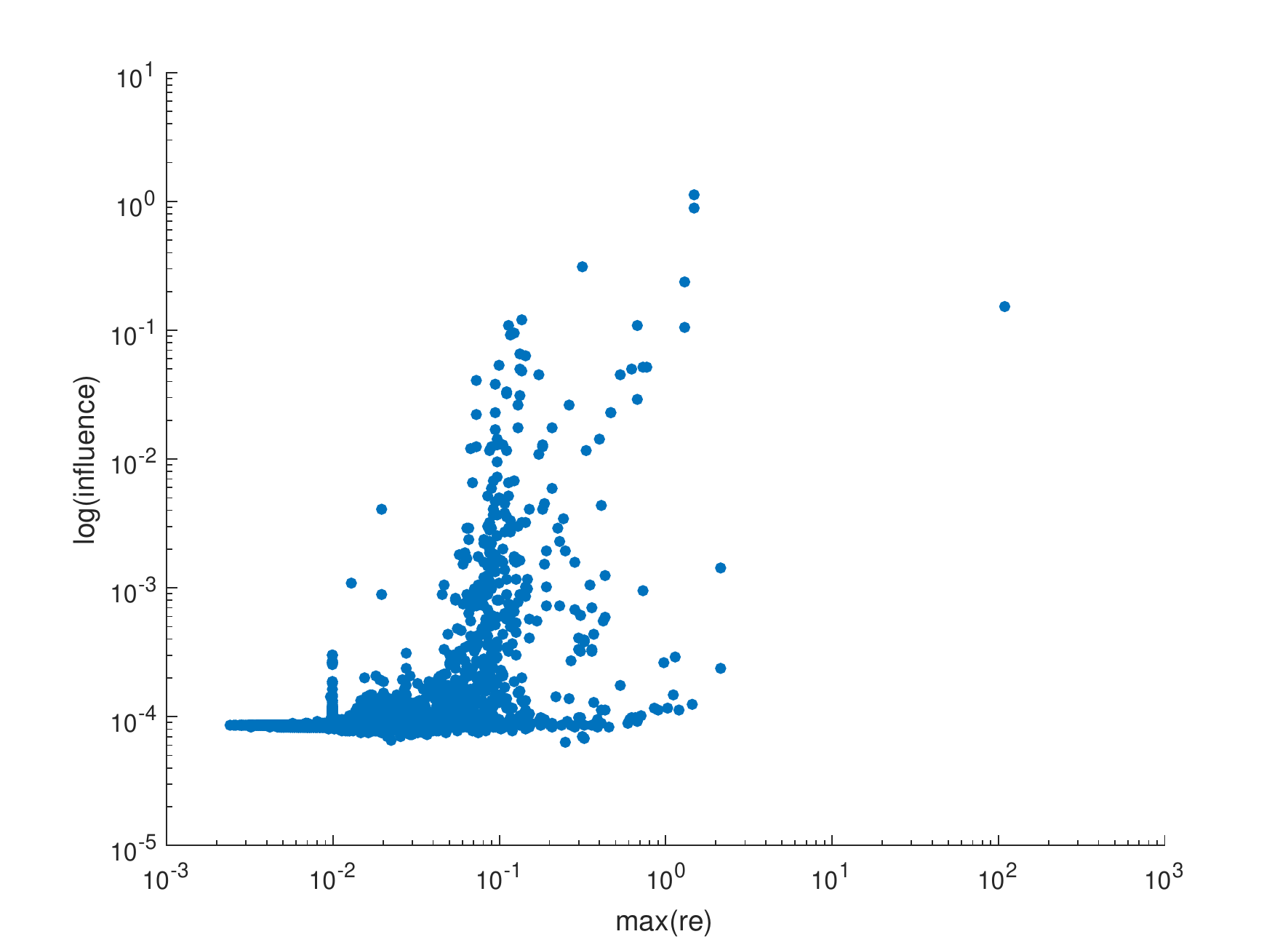}}
\hfill
\subfigure[]{\includegraphics[height=3cm]{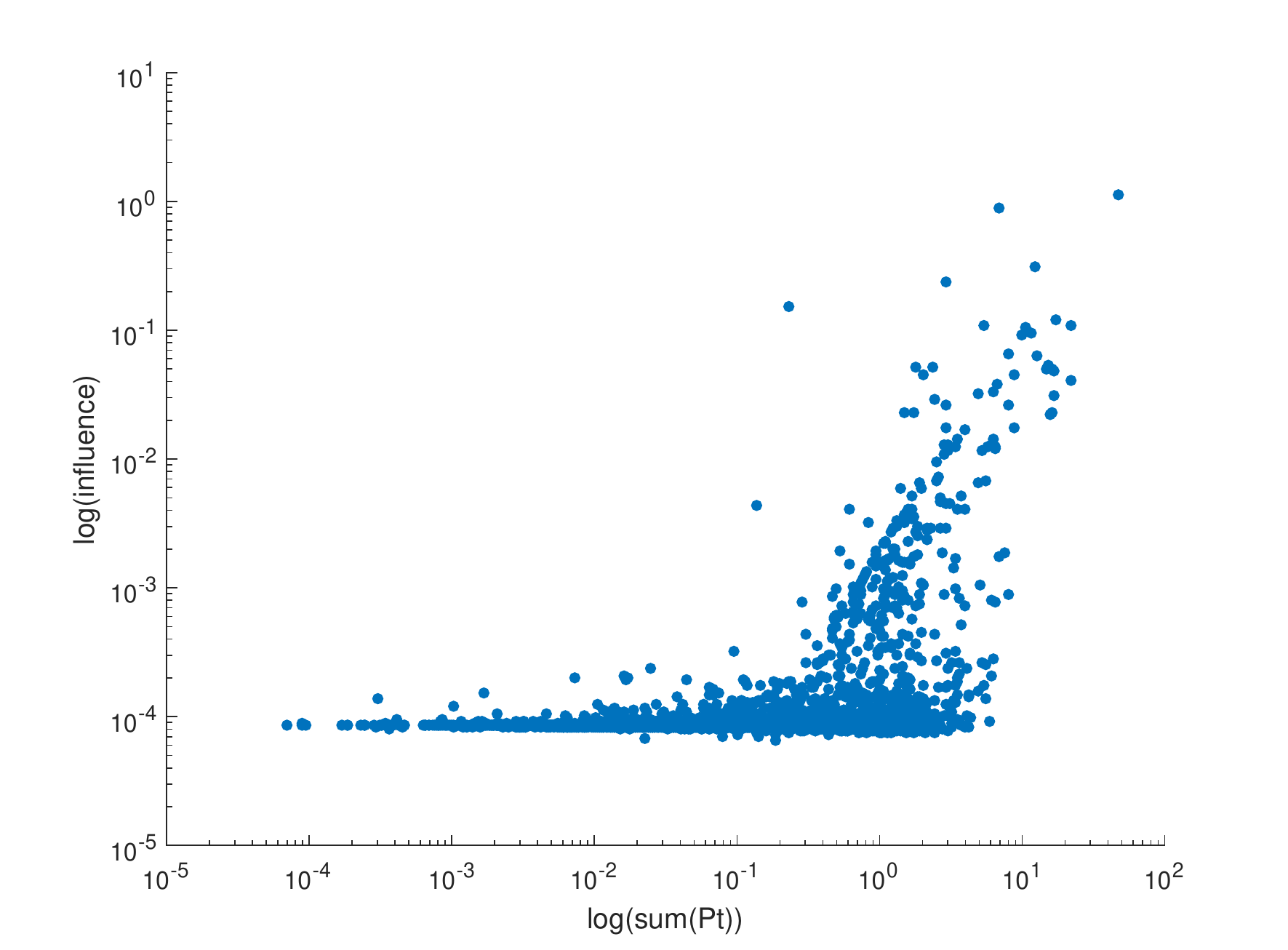}}
\hfill
\subfigure[]{\includegraphics[height=3cm]{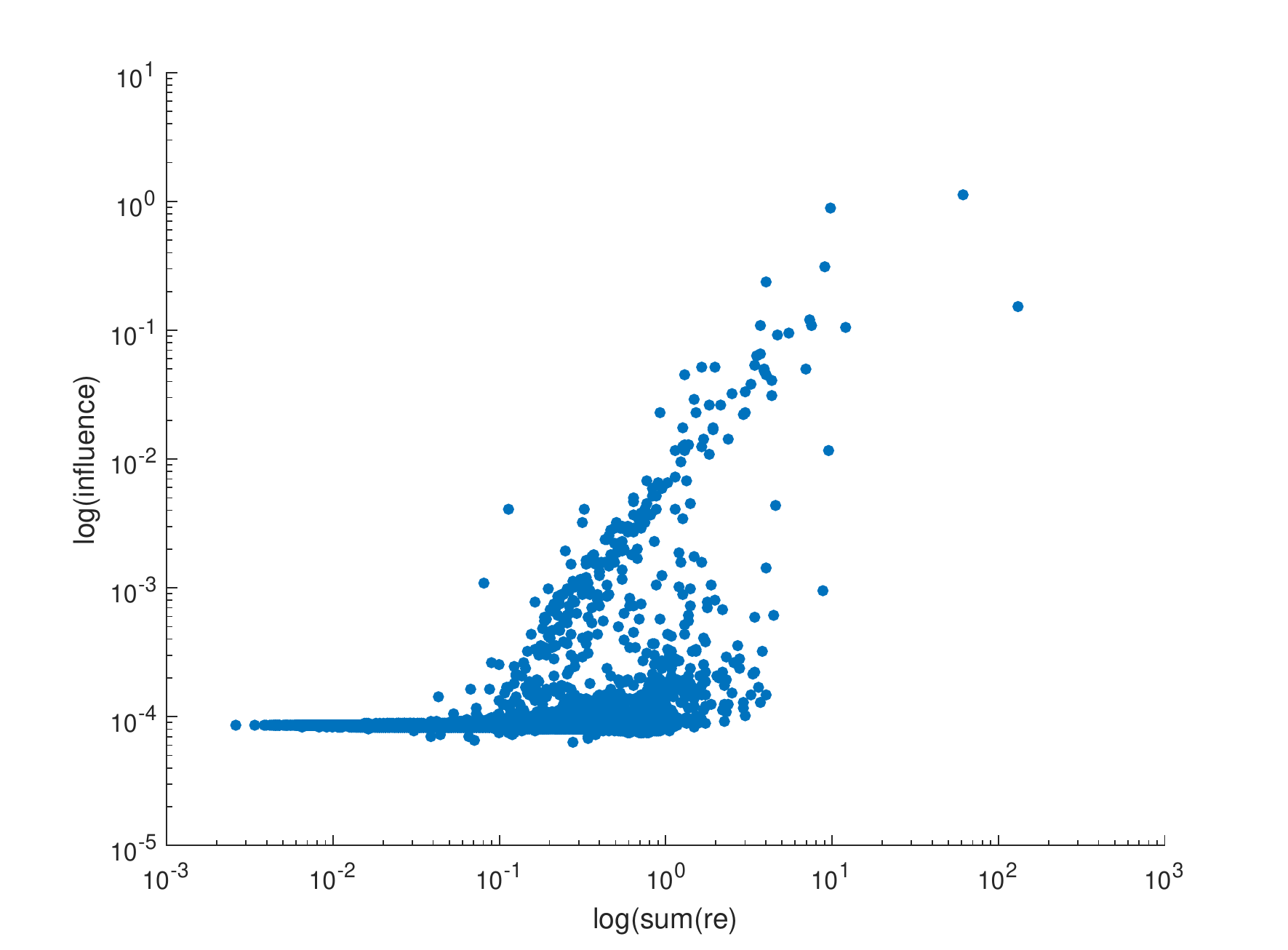}}
\hfill
\caption{Influence of Gulfport superpixels caused by flipping the labels of each superpixel compared with (a) maximum target proportion in each superpixel, (b) maximum residual error in each superpixel, (c) sum of target proportions in each superpixel, and (d) sum of residual errors in each superpixel}
\label{fig:superpixel_influence_Gulfport}
\end{center}
\end{figure}

To understand the correlation between influence of superpixel and proposed approaches, the Gulfport dataset was over-segmented into about seven thousand superpixels having similar sizes using a Normalized Cut method \cite {gillis2012hyperspectral}. Then, the rest of experiment was very similar with single point influence experiment with one difference which is instead of computing the target proportion of all points in each superpixel, we relied on the largest target proportion in each superpixel as the surrogate influence metric.  This process is outlined in Alg. \ref{alg:superpixel}. Figure \ref{fig:superpixel} (a) shows the superpixels of Gulfport dataset. Figure \ref{fig:superpixel_influence_Gulfport} illustrates the plot of log of the influence value versus the log of largest target proportion, residual error, sum of target proportion, and sum of residual error in each superpixel, respectively.  As is shown in Figure \ref{fig:superpixel_influence_Gulfport}, the log of the influence is most correlated with the largest target proportion in each superpixel among all proposed estimation methods. 

\begin{algorithm}

\caption{Superpixel Experiment}
\algsetup{indent=4em}
\begin{algorithmic}[1]  
\STATE Segment input image into $N$ superpixels
\STATE Initialize $\mathbf{E}_{init}$ via VCA, $\mathbf{P}_{init}$ as $\frac{1}{M+1}$
\STATE $\mathbf{E}_{0},\mathbf{P}_0 \gets $ $e$FUMI($L$, $\mathbf{X}$, $\mathbf{E}_{init}$, $\mathbf{P}_{init}$)
\STATE Compute residual error for each data point: $\mathbf{r} \gets \left\| \mathbf{X} - \mathbf{E}_{0}\mathbf{P}_0 \right\|_2^2$, compute max($\mathbf{r}$) in each superpixel, compute sum($\mathbf{r}$) in each superpixel
\STATE Obtain target proportion for each data point, $\mathbf{p}_t$ via unmixing using \eqref{eqn:unmix}, compute max($\mathbf{p}_t$) in each superpixel, compute sum($\mathbf{p}_t$) in each superpixel
\FOR{i=1:$N$} 
\STATE Flip all of the labels for each data point in superpixel $\mathbf{s}_{i}$ to obtain $L^i$
\STATE $\mathbf{E}_{i},\mathbf{P}_{i} \gets $ $e$FUMI($L^i$, $\mathbf{X}$, $\mathbf{E}_{0},\mathbf{P}_{0}$)
\STATE Compute Influence: $\mathbf{I}_{i}$ using \eqref{eqn:Influence}
\STATE Restore the labels to $L$
\ENDFOR

   \RETURN \\
\end{algorithmic} 
\label{alg:superpixel}
\end{algorithm}

\begin{figure}
\begin{center}
\hfill
\subfigure[]{\includegraphics[height=3cm]{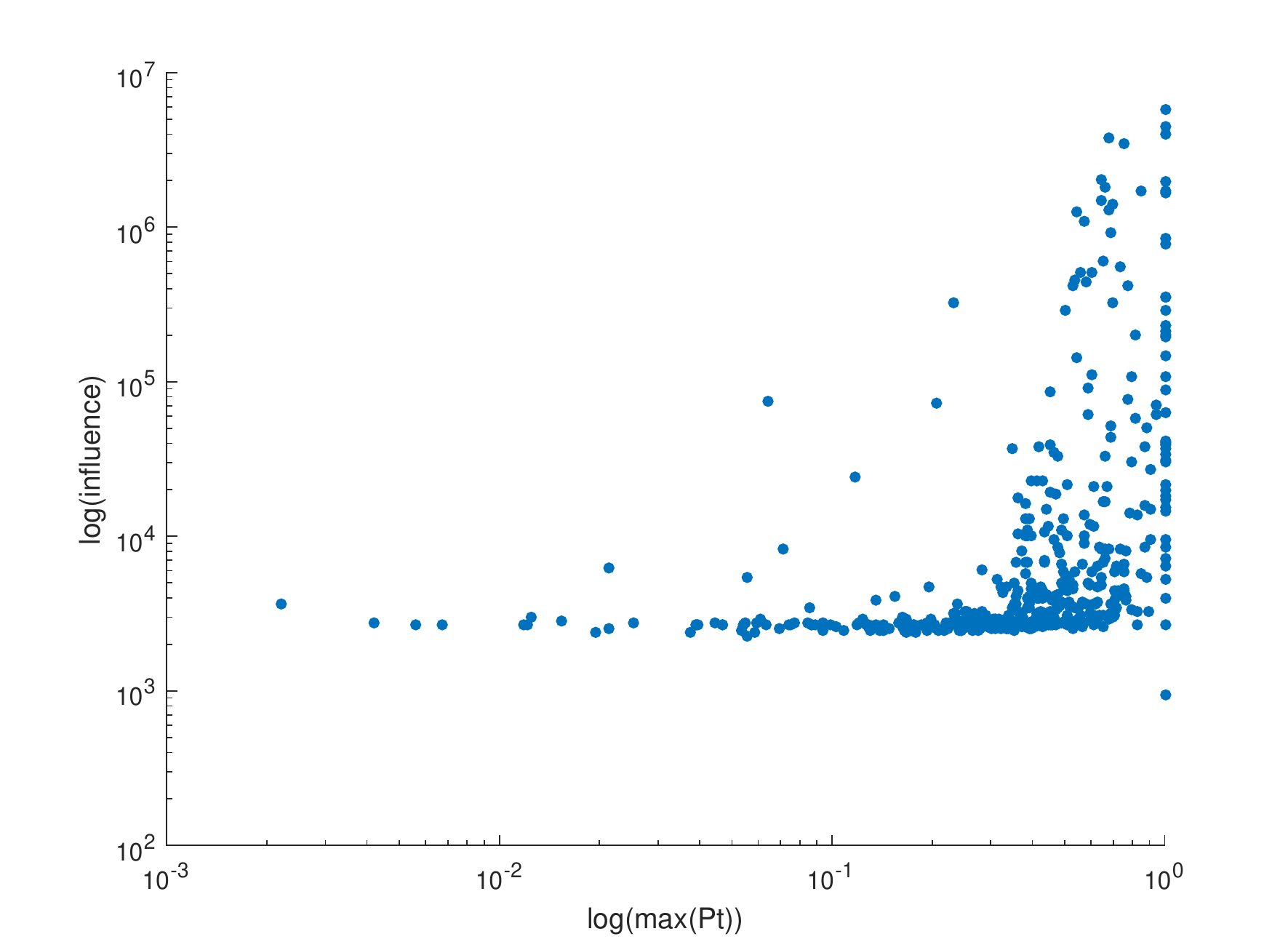}}
\hfill
\subfigure[]{\includegraphics[height=3cm]{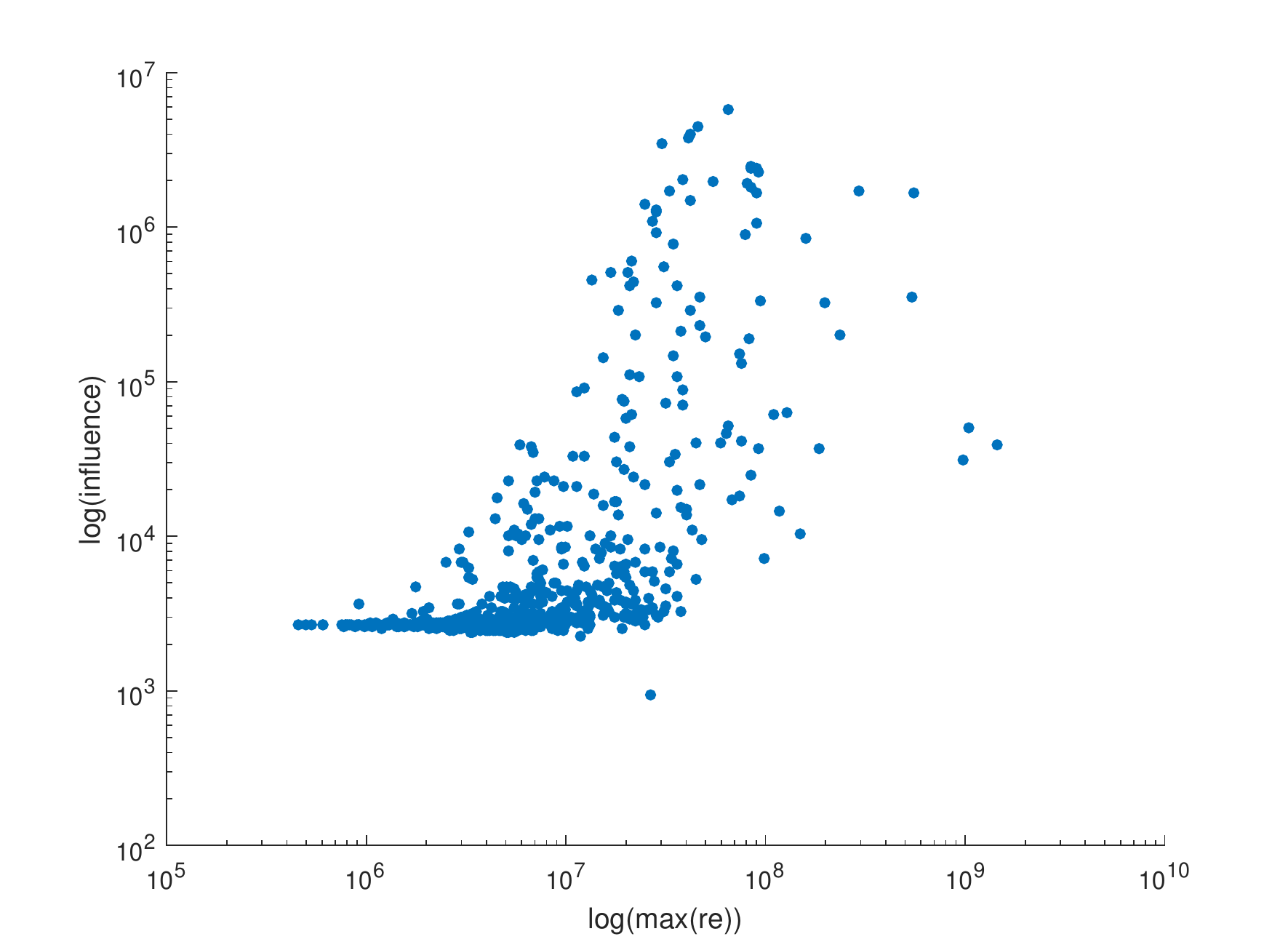}}
\hfill
\subfigure[]{\includegraphics[height=3cm]{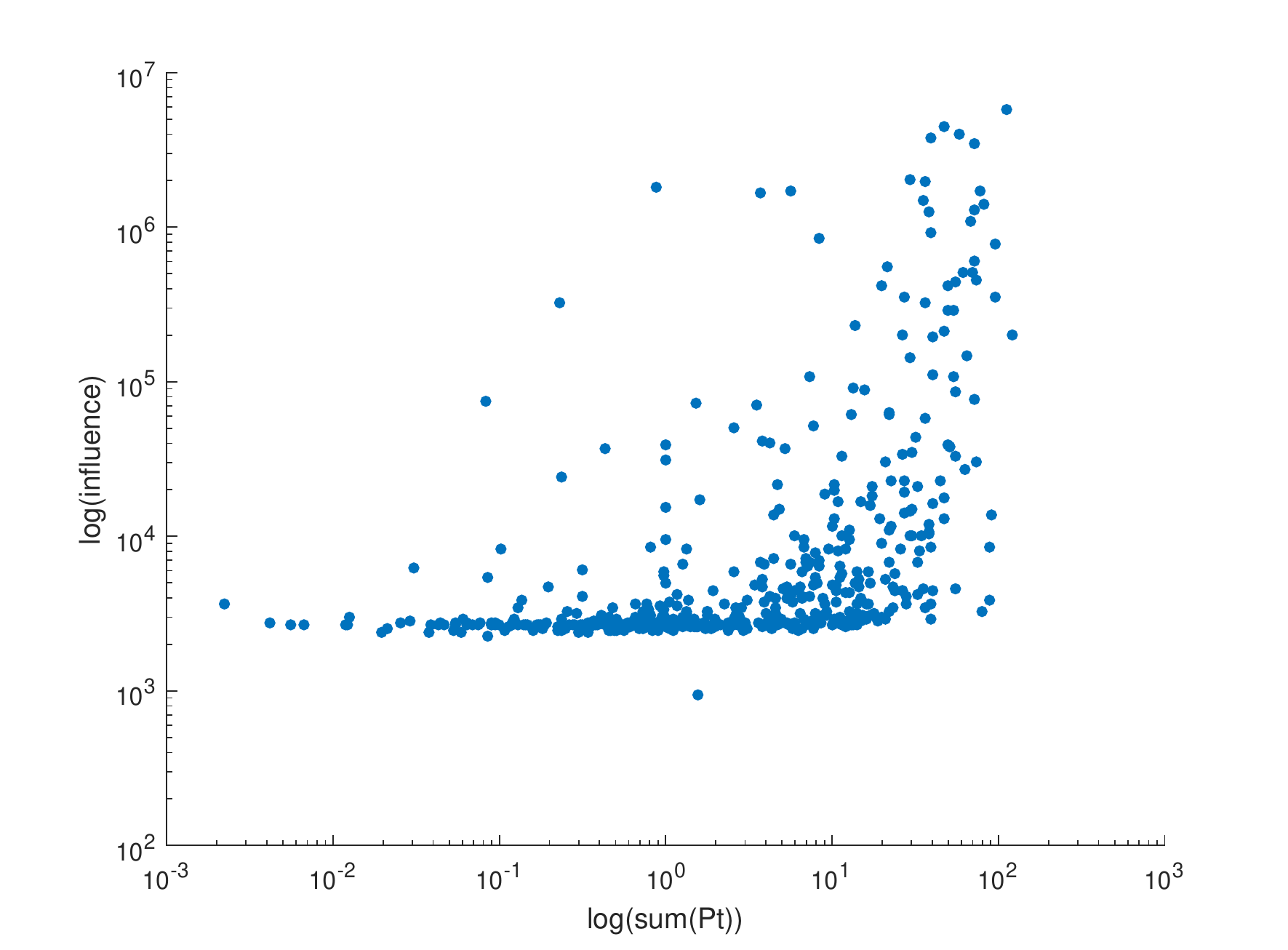}}
\hfill
\subfigure[]{\includegraphics[height=3cm]{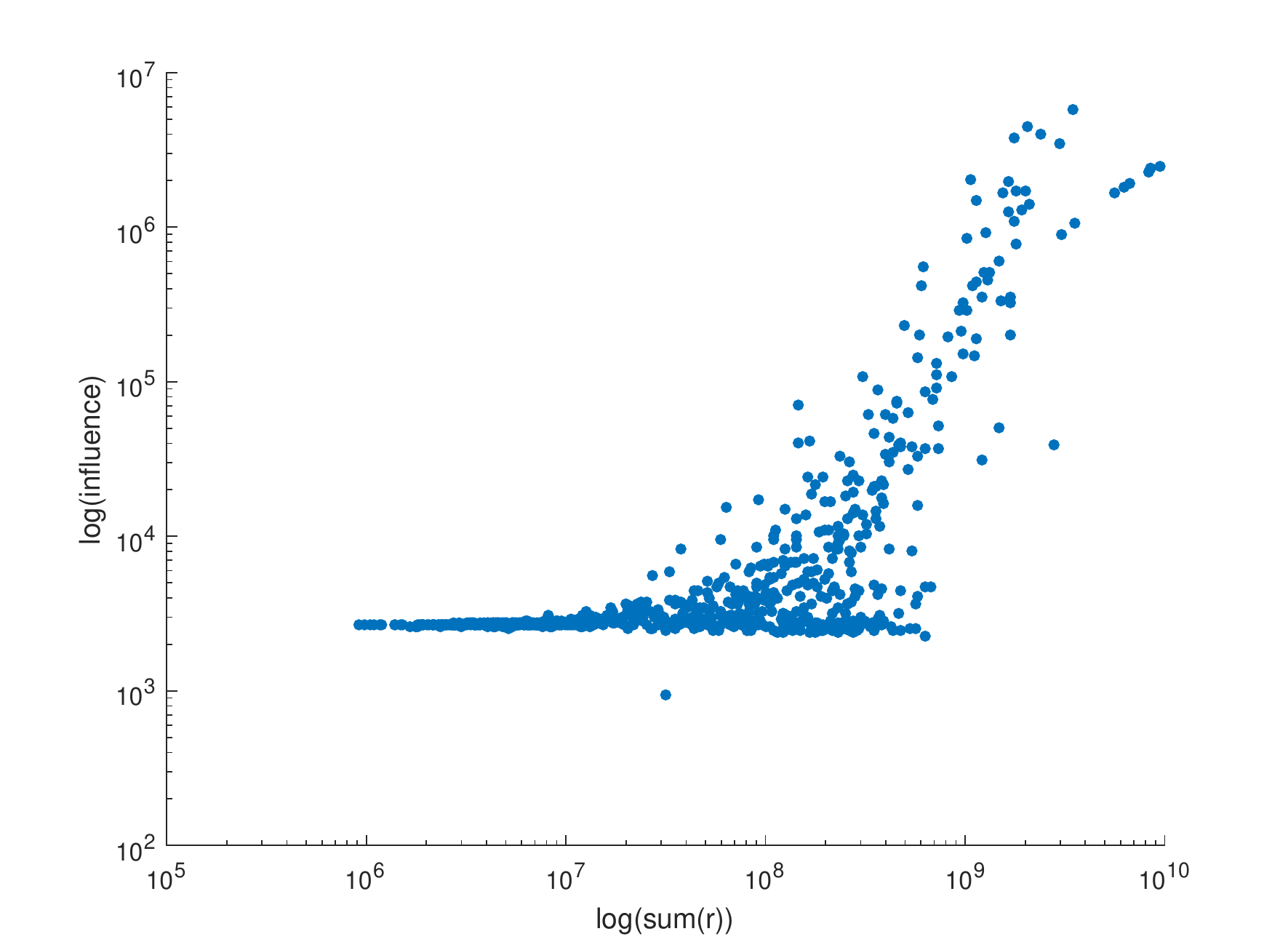}}
\hfill
\caption{Influence of Pavia superpixels caused by flipping the labels of each superpixel compared with (a) maximum target proportion in each superpixel, (b) maximum residual error in each superpixel, (c) sum of target proportions in each superpixel, (d) sum of residual errors in each superpixel}
\label{fig:superpixel_influence_Pavia}
\end{center}
\end{figure}

To further investigate superpixel influence, the same experiment was repeated on the Pavia dataset. Nearly seven hundred segments was generated and shown in \ref{fig:superpixel} (b). The average size of Pavia segments is much larger than those in Gulfport. Figure \ref{fig:superpixel_influence_Pavia} shows the influence value versus all of the proposed metrics for the Pavia data set. The performance of the sum of target proportions and  the sum of the residual errors were found to be more effective than the maximum values.  We believe this to be the case because the sizes of the superpixels on Pavia are much larger than Gulfport so that considering only one maximum point in the superpixel is not comprehensive enough as compared to metrics that take into account all target proportions or residual errors in each superpixel.

\section{Summary}

In this article, we proposed relative influence estimation metrics for the $e$FUMI algorithm.  These provided a mechanism to estimate relative pixel and superpixel influence in $e$FUMI. Generally, the target proportion metric provided better performance as compared to the residual error metric.

\bibliography{report} % bibliography data in report.bib

\begin{thebibliography}{1}

\bibitem{jiao2015functions}
Jiao, C. and Zare, A., ``Functions of multiple instances for learning target
  signatures,'' {\em IEEE Transactions on Geoscience and Remote Sensing}~{\bf
  53}(8),  4670--4686 (2015).

\bibitem{maron1998framework}
Maron, O. and Lozano-P{\'e}rez, T., ``A framework for multiple-instance
  learning,'' {\em Advances in Neural Information Processing Systems} ,
  570--576 (1998).

\bibitem{zare2014extended}
Zare, A. and Jiao, C., ``Extended functions of multiple instances for target
  characterization,'' in [{\em 6th IEEE 2014 Workshop on Hyperspectral Image
  and Signal Processing: Evolution in Remote Sensing
  (WHISPERS)}{\nolinebreak\hspace{0.1em}]},   1--4 (2014).

\bibitem{heinz2001fully}
Heinz, D.~C. and Chang, C.-I., ``Fully constrained least squares linear
  spectral mixture analysis method for material quantification in hyperspectral
  imagery,'' {\em IEEE Transactions on Geoscience and Remote Sensing}~{\bf
  39}(3),  529--545 (2001).

\bibitem{gader2013muufl}
Gader, P., Zare, A., Close, R., Aitken, J., and Tuell, G., ``Muufl gulfport
  hyperspectral and lidar airborne data set,'' {\em Univ. Florida, Gainesville,
  FL, USA, Tech. Rep. REP-2013-570}  (2013).

\bibitem{dopido2013semisupervised}
D{\'o}pido, I., Li, J., Marpu, P.~R., Plaza, A., Bioucas~Dias, J.~M., and
  Benediktsson, J.~A., ``Semisupervised self-learning for hyperspectral image
  classification,'' {\em IEEE Transactions on Geoscience and Remote
  Sensing}~{\bf 51}(7),  4032--4044 (2013).

\bibitem{pavia}
``Hyperspectral remote sensing scenes,'' (2002).
\newblock
  \url{http://www.ehu.eus/ccwintco/index.php?title=Hyperspectral_Remote_Sensing_Scenes}.

\bibitem{nascimento2005vertex}
Nascimento, J.~M. and Dias, J. M.~B., ``Vertex component analysis: A fast
  algorithm to unmix hyperspectral data,'' {\em IEEE Transactions on Geoscience
  and Remote Sensing}~{\bf 43}(4),  898--910 (2005).

\bibitem{gillis2012hyperspectral}
Gillis, D.~B. and Bowles, J.~H., ``Hyperspectral image segmentation using
  spatial-spectral graphs,'' in [{\em SPIE Defense, Security, and
  Sensing}{\nolinebreak\hspace{0.1em}]},   83901Q--83901Q, International
  Society for Optics and Photonics (2012).

\end{thebibliography}
\bibliographystyle{spiebib} % makes bibtex use spiebib.bst

\end{document}